\documentclass[twocolumn,5p, final]{elsarticle}
\biboptions{sort&compress}
\usepackage[dvipsnames]{xcolor}
\usepackage{amsmath,amsfonts, nccmath}
\usepackage{algorithmic}
\usepackage{array}
\usepackage[caption=false,font=footnotesize]{subfig}
\usepackage{textcomp}
\usepackage{stfloats}
\usepackage{url}
\usepackage{verbatim}

\usepackage[commandnameprefix=ifneeded]{changes}

\usepackage[vlined,linesnumbered,ruled,norelsize]{algorithm2e}
\SetKwInOut{Input}{input}\SetKwInOut{Output}{output}
\usepackage[inline]{enumitem}
\usepackage{tabularx, booktabs,ragged2e, multirow}
\usetikzlibrary {positioning, arrows, calc, arrows.meta, patterns, shapes.geometric, bending, shadings, shapes.arrows, decorations.text}
\usepackage{amsthm}
\theoremstyle{plain}
\newtheorem{hypo}{Hypothesis}
\theoremstyle{remark}
\newtheorem{remark}{Remark}

\journal{Robotics and Autonomous Systems}

\begin{document}

\begin{frontmatter}

\title{Human Leading or Following Preferences: Effects on Human Perception of the Robot and the Human-Robot Collaboration}

\author[inst1]{Ali Noormohammadi-Asl\corref{cor1}}
\author[inst1]{Kevin Fan}
\author[inst1]{Stephen L. Smith}
\author[inst1]{Kerstin Dautenhahn}

\affiliation[inst1]{organization={Department of Electrical and Computer Engineering},
            addressline={University of Waterloo, Waterloo, ON}, 
            country={Canada}}

\cortext[cor1]{Corresponding author. \\Email address: ali.asl@uwaterloo.ca}
\tnotetext[]{This research was undertaken, in part, thanks to funding from the Natural Sciences and Engineering Research Council of Canada (NSERC) and the Canada 150 Research Chairs Program.}

\begin{abstract}
Achieving effective and seamless human-robot collaboration requires two key outcomes: enhanced team performance and fostering a positive human perception of both the robot and the collaboration. This paper investigates the capability of the proposed task planning framework to realize these objectives by integrating human leading/following preference\added{s} and performance into its task allocation and scheduling processes. We designed a collaborative scenario wherein the robot autonomously collaborates with participants. The outcomes of the user study indicate that the proactive task planning framework successfully attains the aforementioned goals. We also explore the impact of participants' leadership and followership styles on their collaboration. The results reveal intriguing relationships between these factors\deleted{,} which warrant further investigation in future studies.
\end{abstract}






\end{frontmatter}


\section{Introduction}

Human-robot collaboration provides the opportunity to exploit the complementary abilities of both the human and the robot in a collaborative task. This collaboration has the potential to offset human limitations, given that robots can be advantageous in tasks that require extensive repetitions, precision, speed, physical strength, and stamina. Collaborative robots (cobots) could also reduce the human's cognitive load of tasks. Simultaneously, it leverages \replaced{humans'}{human} superior abilities to adapt and handle uncertainties and address situations that demand human-level intelligence and decision making \cite{paliga2022human, roveda2021human}.  
However, the presence of humans requires thoughtful consideration of human agents in the development of computational architectures for effective human-robot collaboration \cite{el2019cobot, panagou2023scoping}.

 
The robots' ability to adapt to their human teammates is essential for fostering effective, fluent, and long-term collaboration, establishing a high level of human perception of the robot and the collaboration. Human preference, a focus of many recent studies, is a crucial factor that needs consideration in the design of human-robot collaborative scenarios and in robot programming \cite{mangin2022helpful, nemlekar2023transfer, wang2018facilitating, grigore2018preference, wang2021predicting, nemlekar2022towards}. This paper specifically delves into human preferences within the context of task planning and scheduling. The robot's high computational and planning abilities enable it to assume a more significant and leading role in task planning. However, there are situations (e.g., changing environments) that necessitate human agents taking the leading role and planning for the team. Additionally, different human teammates may have varying leading/following preferences.

Although the robot needs to consider the human agent's leading/following preference and adapt its planning accordingly, the human agent's performance may conflict with the overall team goal and deteriorate team performance. This necessitates \added{that} the robot programming \deleted{to} take into account the human agent's preference and performance simultaneously. However, the human agent's preference and performance are not fixed and may change during the collaboration. For example, for a challenging part of a task, a human agent with the overall leading preference \replaced{might}{would} choose to follow the robot, or a human with an overall high performance may get tired at some point and thus start showing poor performance.

The core concept of the proposed framework for involving the human agent's preference and performance in the robot's programming is \replaced{to enable}{enabling} the robot to adjust its planning according to the human agent's preference\added{,} as long as the human preference does not lead to a substantial deviation from the optimal plan or compromise performance beyond acceptable limits.  

\begin{figure}[!t]
\centering
\input{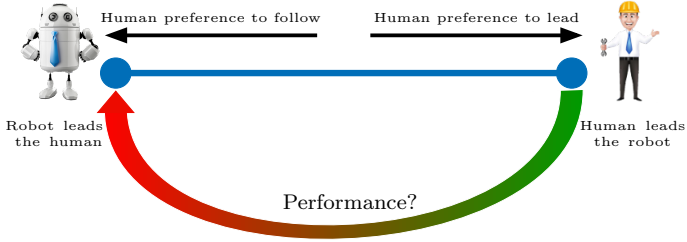}
\caption{Encompassing the entire spectrum of leading/following roles based on human preference and performance}
\label{fig:fol_lead_framework}
\end{figure}

Within this framework, unlike the majority of conventional approaches that concentrate strictly on one extreme of the leading/following spectrum, our approach can cover the entire range, allowing for roles ranging from solely acting as a leader to exclusively functioning as a follower. Fig.~\ref{fig:fol_lead_framework} encapsulates the main idea of this framework. 
This framework requires the robot to continually monitor and, in real-time, gradually adapt to human preferences\replaced{,}{and} performance and their changes. This framework is versatile and applicable to various collaborative scenarios, but our research specifically delves into the realms of task selection/allocation and scheduling within the context of human-robot collaboration.

The fundamental idea of task allocation problems is determining the assignments of tasks to agents in order to achieve the overall team objectives and optimize the team payoff. This has been studied extensively in both multi-robot teams \cite{Khamis2015} and human-robot collaboration (HRC), which incorporates human and robot agents \cite{schmidbauer2023empirical, pupa2022resilient, cheng2021human, lamon2019capability, darvish2018interleaved}. Our work centers on proactive task allocation for single-human \deleted{,} single-robot teams,  in which the human and robot have the agency to select their own tasks and assign tasks to each other. This agency makes the problem different from offline and reactive task allocation problems, where tasks are assigned by a single agent or an external entity like a manager or central controller.

The human agent's agency in selecting and assigning tasks allows them to implement and adjust their leading/following preference. Subsequently, the robot must continually gauge the current human preference and performance in real-time, throughout the collaboration, and adapt its planning accordingly through \added{a} two-step task planning \added{process} at each decision step: first, task allocation, and second, task scheduling. The robot can reassume the leading role in instances of poor human performance, even if the human agent prefers to lead the team. The concept of the robot reassuming the leading role in favor of team performance is rooted in the results of our earlier user study \cite{noormohammadi2021effect}, encompassing three distinct robot strategies: prioritizing the human, prioritizing the robot, and striking a balance between both. This prior work revealed that maintaining a balanced strategy can enhance team performance without compromising the human perception of the collaboration.

Previously, we evaluated the suggested framework by applying it in a simulation environment with a simplified human decision-making model \cite{noormohammadi2022task}, and later on a physical robot, where the experimenter enacted various scenarios of how a human agent might behave \cite{noormohammadi2023adapting}. These scenarios spanned from a high leading preference to a high following preference, as well as from high performance to low performance. This work extends those studies by testing the framework's effectiveness and applicability by recruiting participants to collaborate with the robot in a collaborative scenario, similar to the kitting task. We evaluated participants' perceptions of the robot and collaboration as well as their collaboration strategy and decisions, which are the focus of this paper. We also analyzed the robot's estimations and actions, which are not within the scope of this paper and \replaced{is}{will be} discussed in \replaced{the companion paper}{a future publication} \cite{fetch_robot}.

\added{While the user study in this paper focuses on the kitting task scenario, the task planning framework and algorithm can be adapted to various collaborative tasks. For instance, in a co-assembly task discussed in \mbox{\cite{schmidbauer2023empirical}}, using our framework the robot could design an efficient plan and allocate tasks based on its computational abilities and understanding of its limitations, while also incorporating human preferences and performance. The robot must balance task efficiency with human preferences, learning whether the human prefers to assign tasks or be assigned tasks, as well as their preferred types of tasks. As long as human preferences do not significantly compromise team efficiency, the robot can accommodate them. Additionally, the framework can adapt to changing human preferences, such as a human shifting from assigning tasks to preferring the robot to allocate tasks when the task becomes challenging or they are fatigued.}

Given the intended objective of the proposed framework -- to empower the robot to adapt in real-time to the preferences and performance of its human teammate while \replaced{maintaining}{retaining} a high level of human perception regarding the robot and collaboration  -- we explore the following research questions:
\begin{itemize}

    \item RQ1: To what extent can this collaboration and task planning framework improve the team performance and reduce participants' perceived task load?
    \item RQ2: How do the human agent's perception of the robot and the collaboration affect their collaboration with the robot and vice versa?
    \item RQ3: How does this collaboration influence the human agent's actions and decision making?
    \item RQ4: How does the task difficulty influence participants' actions and collaboration with the robot?

\end{itemize}

\subsection{Contributions}
The contributions of this work are sevenfold:
\begin{enumerate}
    \item Designing a collaborative scenario involving autonomous pick-and-place tasks for the robot and implementing the task planning framework along with the estimation methods for its execution.
    
    \item \added{Implementing our proposed framework that integrates human preferences and performance into the robot's task planning and scheduling, covering the entire spectrum of leading and following roles.}

    \item \added{Implementing a two-step task planning process that includes task allocation and task scheduling, enabling the robot to dynamically estimate human preferences and performance, adapting its leading/following role, and reassuming the leading role in instances of poor human performance or changing preference (e.g., due to task difficulty) while maintaining optimal collaboration efficiency and effectiveness.}

    \item Implementing dynamic task updates that take into account task states, errors made by the human agent, and subsequent actions by the robot to rectify them.

    \item Studying the effects of the cobot's adaptive task planning on participants' perception of the robot and collaboration, e.g., trust, satisfaction, robot intelligence, and workload.
    \item Studying the effect of participants' preferences and attitude toward collaboration on the cobot's planning.
    \item Studying the effects of participants' leadership style and initial trust in the robot on their collaboration and perception of the robot.
\end{enumerate}

The earlier versions of this research were presented in conference papers~\cite{noormohammadi2022task, noormohammadi2023adapting}. The paper~\cite{noormohammadi2022task} primarily concentrated on designing a robot planning framework and evaluating it in a simulation environment for a collaborative scenario with a simplified human decision-making model. In contrast, this paper extends the framework's application to a more complex collaborative scenario\deleted{,} involving an actual robot collaborating with recruited participants.

The paper~\cite{noormohammadi2023adapting} introduced the initial implementation of this collaborative scenario and assessed the planning framework's performance across four different scenarios \replaced{enacted}{conducted} by the experimenter. However, the present paper unveils the final version of the experimental setup and planning framework. Importantly, it evaluates the effectiveness of the proposed planning framework for an autonomous robot collaborating individually with each of the 48 participants \replaced{on}{, involving} 144 tasks. This paper focuses on evaluating participants' perceptions of the robot, collaboration, and tasks, examining their effects on actions and collaboration. In \cite{fetch_robot}, the companion paper explores the framework's ability to enable the robot to plan and adapt to participants' preferences and performance.
\subsection{Organization}
The rest of this paper is organized as follows: Section 2 reviews pertinent literature and formulates the research hypotheses. Section 3 outlines the study's design, including both objective and subjective measures, and details the study procedure. In Section 4, we analyze the results, emphasizing participants' perceptions of the robot, tasks, collaboration, and how these factors impact their decision-making. Section 5 concludes the paper, addressing its limitations and suggesting potential avenues for future research.

\section{Related Work \& Hypotheses Development}

In this section, we initially conduct a review of pertinent literature focusing on the human perception of the robot and collaboration in terms of trust, self-confidence, reliance, workload, and helpfulness, and develop our hypotheses based on the insights gathered from the existing literature. We also explore the human perception of control and the various leadership and followership styles. Finally, we briefly review research concerning task allocation and adaptation to human preferences.

\subsection{Human Perception of the Robot \& Collaboration}
\subsubsection{Trust, self-confidence \& reliance}
Trust is a complex and multifaceted concept in human-human, human-technology, and, more specifically, in human-robot interactions (HRI). 
In the context of human-robot trust, contributing factors to human trust in robots can be categorized into three groups \cite{sanders2011model}: 
\begin{itemize}
    \item robot characteristics and design elements (e.g., reliability)
    \item human characteristics (e.g., self-confidence)
    \item environmental-based factors (e.g., task type)
\end{itemize}
Typically, in most human-robot collaboration scenarios, it is assumed that humans and robots share common goals, and the human agent is aware of the collaborative context of the interaction, which prevents them from considering deception. Sanders et al. mainly focussed on the first two factors mentioned above and investigated how robot performance can affect human trust \cite{sanders2011model}.

 Trust is a topic of interest in HRI and HRC and has been studied extensively. Some studies focus on how to model and measure trust and its dynamics and, in some cases, use it to adjust the robot's behavior or improve human distrust or overtrust in the robot \cite{khavas2020modeling, soh2020multi, chen2018planning}.   Some other studies, including this work, use objective or subjective measures (e.g., questionnaires) to evaluate human trust in robots before or after interaction with a robot. In our study, using Muir’s questionnaire \cite{muir1996trust}, we are interested in how participants' trust in the robot evolves during the sessions over time. We established the following hypothesis:

\begin{hypo}\label{hypo1a}
Participants' trust in the cobot, measured before and during the collaborative tasks, will improve during the session while engaged in different tasks.
\end{hypo}

Furthermore, it is well-established that human reliance on an autonomous system is intertwined with their trust in the system and their self-confidence (i.e., their trust in their capability to accomplish the task). Studying and modeling this relationship has been the focus of many studies. The ``confidence vs. trust" hypothesis is a well-accepted concept and serves as the foundation for many proposed models \cite{lee1994trust}. It posits that a person's reliance on an automation system is influenced by their relative trust, which refers to the difference between their trust in automation and their self-confidence. That is, if an individual's trust in the system exceeds their self-confidence, they will opt to rely on the system; otherwise, they will rely on their own decisions and perform the task on their own. However, the findings of some studies contradict this hypothesis \cite{wiczorek2019effects}. \replaced{The authors}{Authors} in \cite{williams2023computational} have proposed a computational model that outperforms the ``confidence vs. trust" hypothesis and signifies nuances between trust and self-confidence in predicting human reliance on automation. It is also noteworthy that in a collaborative scenario, the goal is to exploit human and robot abilities, not bypassing one by employing the other. This point may affect the full applicability of the ``confidence vs. trust" hypothesis in human-robot collaboration. 
 
\added{In our study, we measured self-confidence by asking participants a question at different stages of the user study.} We state the associated hypotheses as follows:

\begin{hypo}\label{hypo2A}
Participants' self-confidence will improve during the collaborative tasks over time as they collaborate more with the robot.
\end{hypo}


\begin{hypo} \label{hypo2B}
For demanding tasks, participants will trust the robot more than their own ability and will rely more on the robot.
\end{hypo}

\subsubsection{Perceived workload}
There are three fundamental measures for evaluating workload: physiological (e.g., heart rate), performance-based (e.g., time), and perceptual (subjective) \cite{miller2001workload}. In this paper, we consider performance-based and subjective measures. For the subjective measure, we use NASA-TLX, which defines the workload experience in terms of ``the sources of loading imposed by different tasks" \cite{hart1988development}. We seek to study whether collaborating with the robot could positively influence participants' perceived workload, and derive the following hypothesis. 

\begin{hypo}\label{hypo4}
Participants will experience less workload when collaborating with the cobot rather than working alone, in terms of the dimensions in the NASA task load index, including:
\begin{enumerate*}[label=(\alph*)]
\item mental demand
\item physical demand
\item performance
\item effort
\item frustration
\end{enumerate*}. However, the temporal demand will be similar,  as there \replaced{is}{was} no time limit for completing tasks.
\end{hypo}

\subsubsection{Helpfulness}
In \cite{freedman2020helpfulness}\added{,} helpfulness is defined as ``with honest intentions, trying to play a positive role with the task at hand." In this study, we aim to investigate how humans' expected helpfulness from the robot, which refers to their anticipation of the robot's ability to provide effective assistance and enhance their overall collaboration experience, changes during collaboration. \added{To measure participants' expected helpfulness of the robot, we asked them to answer a question about it at different stages of the user study.}

\begin{hypo}\label{hypo3}
Participants' expected helpfulness of the cobot will increase over time as they collaborate more with the cobot.
\end{hypo}

\subsection{Being in or under control}

While the concept of human-robot collaboration is a relatively recent subject, conversations surrounding human automation have endured for many years. The rise of automation systems has gradually replaced humans in various control systems, \replaced{ranging}{spanning} from simple to complex setups. Different factors, such as human preferences and abilities, contribute to determining the autonomy levels within these systems. These can range from manual control, where human operators actively control, to semi-autonomous and fully autonomous systems, where human operators take on more supervisory roles.

The perception of control \replaced{in}{by} humans is a crucial area of research, spanning psychology, social science, and human-computer (robot) interaction. In the late 1990s, researchers like Ben Shneiderman and Pattie Maes discussed human-computer interface design from distinct perspectives \cite{shneiderman1997direct}. Shneiderman emphasized user-controlled direct manipulation, highlighting human control, predictability, and responsibility. In contrast, Maes advocated for long-lived, proactive, \added{and} adaptive software agents aligned with individual user habits and preferences. As of 2023, traditional input methods like mice and keyboards remain predominant in computer interactions, suggesting a prevailing inclination toward control and responsibility when dealing with technology. This underscores the significance of design decisions and user anticipations when incorporating intelligent technologies into human-computer interactions, comprehending users' behaviors and control perceptions across diverse levels of agency and automation \cite{dove2017ux, heer2019agency}. When extending this concept to human-robot interaction, a preference is likely for humans to maintain control over robots.

The perception of robot control by humans can be influenced by factors such as human characteristics (e.g., locus of control), trust in robots, self-confidence, task difficulty, task criticality, human abilities, and robot abilities. In \cite{pasparakis2021control}, the issue of being in control or under control was explored in a warehouse order-picking scenario, demonstrating higher productivity when the human leads and higher accuracy when the human follows the robot. In contrast, ``6 River Systems," a company actively engaged in collaborative robotics and warehouse automation, contends that allowing associates to have self-control over their pace leads to decreased speed and efficiency \cite{6river}. In response to this challenge, they introduced a ``system-directed picking" approach, wherein robots guide associates through their tasks. Another instance in the industrial domain is the robotic picking assistive system implemented in DHL warehouses, where the robot (supplied by Locus Robotics) directs human workers to the designated pickup location \cite{dhl}. In these setups, the robot is tasked with planning the tasks, while humans are responsible for picking up items.

The findings from a user study in \cite{chanseau2019does}, in the context of household robots, indicate that participants felt a greater sense of control when the robot operated in semi-autonomous mode. They observed a preference among participants for reduced autonomy in critical tasks (e.g., scheduling a doctor's appointment). However, when it came to a robot transporting biscuits from the kitchen to the living room, contrary to their stated preference, participants leaned towards the autonomous robot condition. The authors attributed this shift to the perceived advantages of higher physical efficiency and comfort. It is worth noting that a more comprehensive analysis could be conducted if the researchers had also inquired about participants' trust in the robot's ability to move autonomously. Furthermore, in \cite{roy2019automation}, the authors explore the connection between automation accuracy and task controllability, emphasizing its impact on user satisfaction. Their study underscores a strong preference for manual control over automation. \added{The user study in \cite{schmidbauer2023empirical} also shows that, in a co-assembly task, participants preferred to retain decision-making authority and take over cognitive tasks.}

In our work, based on the collaborative and cognitively critical  nature of the scenario designed (due to a penalty for mistakes), we state the next hypothesis of this paper as follows:

\begin{hypo}\label{hypo5}
Overall, participants will prefer retaining control over the robot and being a leader.
\end{hypo}

\subsection{Leadership and Followership Styles}
In this study, we are also interested in how participants' general leadership and followership styles can affect their collaboration with the robot.

\textbf{Leadership Styles \cite{northouse2014introduction}:}
Leadership style pertains to the behaviors leaders demonstrate, encompassing their actions and interactions with followers. It characterizes what leaders do and how they behave in diverse circumstances. Leadership style is intricately linked to an individual's personal leadership philosophy. Commonly observed leadership styles include \cite{northouse2014introduction}: \begin{enumerate*}
    \item authoritarian
    \item democratic
    \item laissez-faire
\end{enumerate*}
\begin{itemize}
    \item \textbf{authoritarian --} involves leaders perceiving followers as requiring guidance, exerting control over their actions, emphasizing their authority, and directing communication toward themselves rather than fostering interaction among group members.
    
    \item \textbf{democratic --} involves leaders treating followers as capable collaborators, guiding rather than controlling, and fostering open, inclusive, and supportive communication. They prioritize helping followers achieve personal goals, provide guidance and suggestions without giving orders, and deliver objective praise and criticism in evaluations.
    
    \item \textbf{laissez-faire --}   is a ``hands-off, let it ride" approach where leaders exert minimal influence and allow followers significant freedom. They do not control or guide followers' activities and may serve as temporary or interim leaders with limited authority or interest in shaping outcomes.
\end{itemize}

It is worth emphasizing that these leadership styles are not separate categories; they can overlap and coexist. Leaders may adopt different styles depending on the situation. For example, a leader might exhibit an authoritarian approach in certain contexts while adopting a democratic style in others, showcasing a dynamic and multifaceted approach to leadership. Northouse proposed the leadership style questionnaire in \cite{northouse2014introduction} to identify participants' tendency to the leadership styles. 

\textbf{Followership Styles \cite{kelley1992power}:}
Kelley (1992) identified five followership styles based on two key dimensions: engagement and critical thinking. Engagement ranges from passive (waiting for direction) to active (taking the initiative to participate actively in tasks). Critical thinking spans from dependent uncritical thinking (accepting information without evaluation or questioning) to independent critical thinking (evaluating and analyzing information to identify consequences and opportunities). These five follower styles are:
\begin{itemize}
    \item \textbf{Exemplary}: These followers show high levels of active engagement and independent critical thinking, thinking for themselves, challenging leaders constructively, proactively supporting organizational goals, and assuming extra responsibilities.

    \item \textbf{Conformist:} Conformist followers are actively engaged but tend to be dependent uncritical thinkers, enthusiastically following leader directions without questioning.

    \item \textbf{Passive:} Passive followers have low levels of both engagement and critical thinking, needing constant direction and acting only after explicit instructions.

    \item \textbf{Alienated:} Alienated followers are independent critical thinkers but lack engagement. They are skeptical, may oppose management, and see themselves as mavericks.

    \item \textbf{Pragmatist:} Pragmatists have a moderate level of both engagement and critical thinking. They are uncommitted, prefer maintaining the status quo, and act cautiously, often waiting for crises before taking action.
\end{itemize}
Kelley also designed the questionnaire in \cite{kelley1992power} for determining followership styles.

\subsection{Task Allocation and Adaptation}
Task allocation in human-robot collaboration (HRC) involves assigning tasks to human and robot agents. This can be done offline, relying on prior knowledge to determine task suitability \cite{schmidbauer2020adaptive, muller2016process, tsarouchi2017human, michalos2018method, lamon2019capability, lee2022task}, or online, allowing for flexibility and reactive replanning. \added{
Suitability can be assessed by evaluating how well agents' abilities and limitations align with the tasks' demands and restrictions. In offline methods, once the suitability of agents for each task has been assessed, task allocation can be performed by experts \cite{schmidbauer2020adaptive, muller2016process}. Alternatively, simulation approaches can be used to explore various efficient allocations \cite{tsarouchi2017human, michalos2018method}, or mathematical optimization can be employed to determine an optimal solution \cite{lamon2019capability, lee2022task}.}

\added{In many real scenarios, however, it is not guaranteed that the team can completely follow the offline plan during the execution phase. The first group of online task allocation methods addresses this limitation by enabling the task allocation method to replan and find a new plan that suits the evolving conditions \cite{faroni2023optimal, cheng2021human, pupa2022resilient, alirezazadeh2022dynamic}.}

\added{Another group of online approaches grants human agents more agency to make their own decisions while employing robots in a supportive role by adjusting to the actions and preferences of the human agents \cite{gorur2023fabric, fiore2016planning, nemlekar2021two}. These approaches require incorporating human preferences into task allocation to achieve effective collaboration. Examples of human preferences may include how to assemble different parts \cite{nemlekar2021two}, the types of tasks they prefer to do \cite{gombolay2017computational}, or whether they prefer to command or take a passive role \cite{fiore2016planning, schmidbauer2023empirical}. Studies have demonstrated the positive effects of considering human preferences in planning \cite{tausch2022best, gombolay2017computational}.}

\added{The adaptation of robots to humans has been widely studied in various areas of human-robot interaction, such as designing controllers for physical human-robot interaction \cite{roveda2023optimal, maccarini2022preference}, considering human ergonomic preferences in collaborative tasks \cite{falerni2024framework}, and specifying traffic rules and constraints for robot movement on maps \cite{wilde2020improving}. Robot adaptation can be achieved through preprogrammed instructions by experts \cite{akgun2012keyframe, huang2019synthesizing}, autonomous learning and adaptation capabilities, or a combination of both \cite{unhelkar2020effective, bajcsy2018learning}. Learning-based adaptation approaches employ supervised or unsupervised learning algorithms \cite{sadigh2017active, ayoub2023real, huang2021meta, nemlekar2021two, reddy2022first}, with the former involving identifying factors influencing human behavior and enriching behavioral data through expert annotation or surveys. The latter enables robots to learn human preferences by observing their behavior without explicit annotation. Some research also employs a combination of learning techniques and expert knowledge \cite{unhelkar2020effective, bajcsy2018learning}.}

\added{However, many of these studies focus on the robot adapting to the human agent, while mutual adaptation, where both human and robot adjust based on each other's actions and feedback, is also vital in human-robot collaboration, as demonstrated in the studies exploring mutual adaptation between humans and robots \cite{nikolaidis2017human}. What has been overlooked in these studies is assessing how these preferences align with overall team performance.}

\deleted{Online approaches can either find an optimal allocation \mbox{\cite{faroni2023optimal, cheng2021human, pupa2022resilient, alirezazadeh2022dynamic}} or involve human decisions, with robot agents adapting to human preferences \mbox{\cite{gorur2023fabric, fiore2016planning, nemlekar2021two}}. Incorporating human preferences into task allocation is essential for enhancing the perception of the robot, but this aspect is often underestimated. While some studies have considered human preferences during offline planning, there is a need to assess how these preferences align with overall team performance.}

\deleted{In the context of task allocation, the ability of a robot to adapt is crucial for effective collaboration with humans \mbox{\cite{tausch2022best, gombolay2017computational, fiore2016planning}}. Robot adaptation can be achieved through preprogrammed instruction by experts \mbox{\cite{akgun2012keyframe, huang2019synthesizing}}, equipping the robot with autonomous learning and adaptation capabilities, or a combination of both \mbox{\cite{unhelkar2020effective, bajcsy2018learning}}.  Learning-based adaptation approaches employ supervised or unsupervised learning algorithms \mbox{\cite{ayoub2023real, huang2021meta, nemlekar2021two,reddy2022first}}, with the latter involving identifying factors influencing human behavior and enriching behavioral data through expert annotation or surveys. The former enables robots to learn human preferences by observing their behavior without explicit annotation. However, many of these studies focus on the robot adapting to the human agent, while mutual adaptation, where both human and robot adjust based on each other's actions and feedback, is also vital in human-robot collaboration, as demonstrated in the studies exploring mutual adaptation between humans and robots \mbox{\cite{nikolaidis2017human}}.}

\section{User Study: Setup \& Methodology}
This section provides a detailed explanation of the setup, procedure, and measures employed in this user study.

\subsection{User Study Setup}

The design of the user study scenario was shaped by three pivotal factors: 
\begin{enumerate}
\item \textbf{Collaboration:} Focusing on the planning abilities of both the human and robot, leveraging the cobot's advanced memory capacity and the human's quicker pace,
\item \textbf{Leading/following preference:} Providing the human with the agency to adjust their role as either a leader or follower,
\item \textbf{Performance:} Introducing a task that imposes cognitive load and penalizes errors.
\end{enumerate}

\subsubsection{Setup}
\begin{figure}
    \centering
    \includegraphics[width= \linewidth]{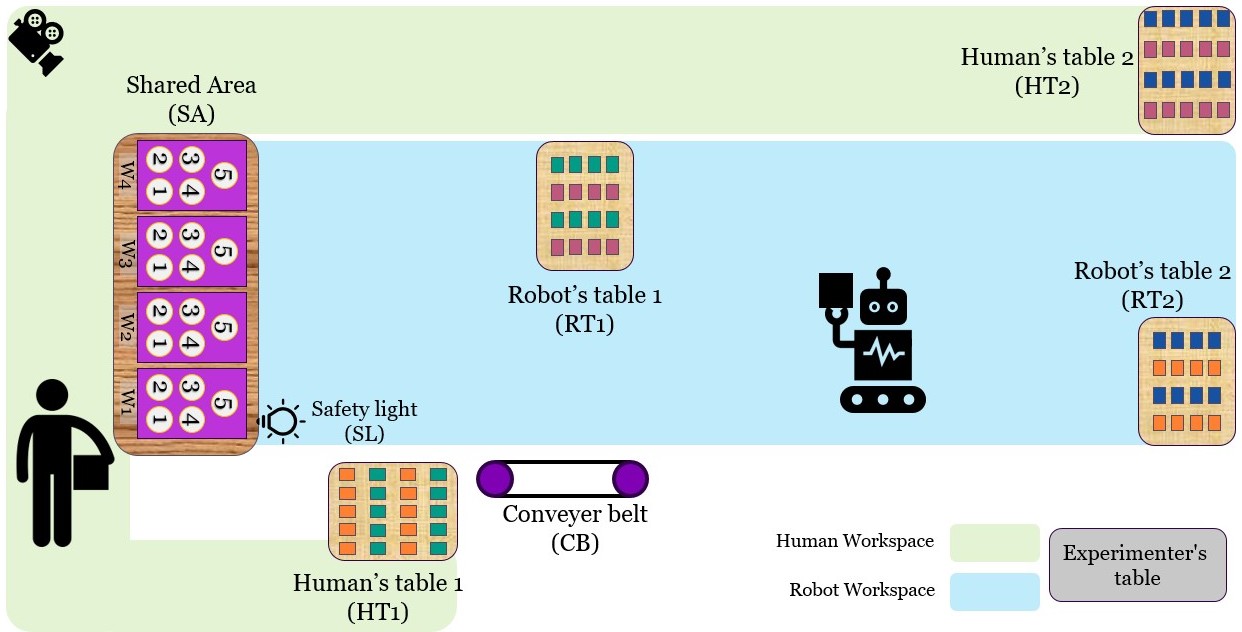}
    \caption{The layout of the experimental setup includes two tables in the robot's workspace and two tables in the human agent's workspace. Additionally, there is a shared area (table) where both agents need to arrange blocks. A conveyor belt beside human table 1 allows the robot to return a block to the human agent. }
    \label{fig:exp_env}

\end{figure}

\begin{figure}
    \centering
    \includegraphics[ width=\linewidth]{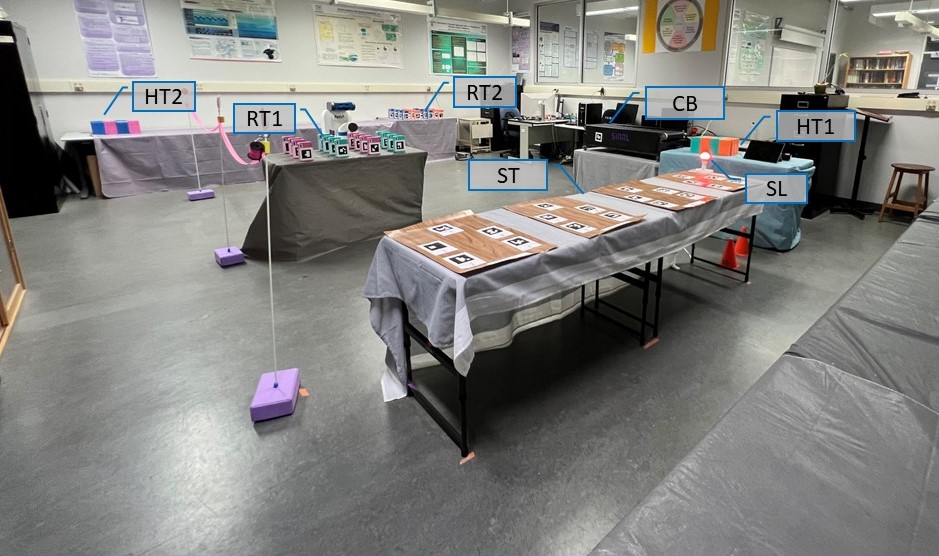}
    \caption{An overview of the experimental setting captured from the perspective of the camera illustrated in Fig.~\ref{fig:exp_env}}
    \label{fig:exp_env_real}
\end{figure}

In this specific scenario, the human-cobot team collaborates to select colored blocks from tables, adhering to a specified pattern, and arrange them within a shared area with a predetermined number of spots. The designed collaborative scenario closely resembles the kitting task. The kitting task involves selecting and collecting all necessary components for assembling a particular final product \cite{krueger2019testing}.  Additional details about the setup are visualized in Fig.\ref{fig:exp_env} and Fig.\ref{fig:exp_env_real}. The camera's placement in Fig.\ref{fig:exp_env} corresponds to the viewpoint from which the image in Fig.\ref{fig:exp_env_real} was captured. Additionally, Fig.~\ref{fig:exp_env} illustrates the location of the experimenter's table, indicating the position of both the computer and the experimenter themselves.

\begin{figure} 
    \centering
  \subfloat[Pattern A\textsubscript{1} \label{fig:pattern_a1}]{%
       \includegraphics[width=0.45\linewidth]{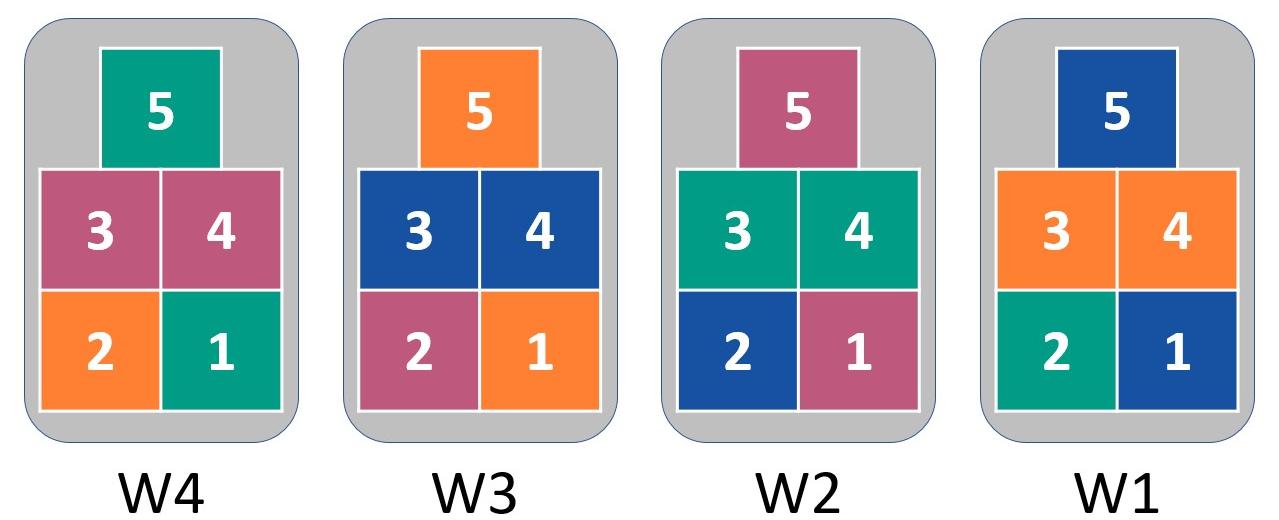}} 
    \hfill
  \subfloat[Pattern A\textsubscript{2} ]{%
        \includegraphics[width=0.45\linewidth]{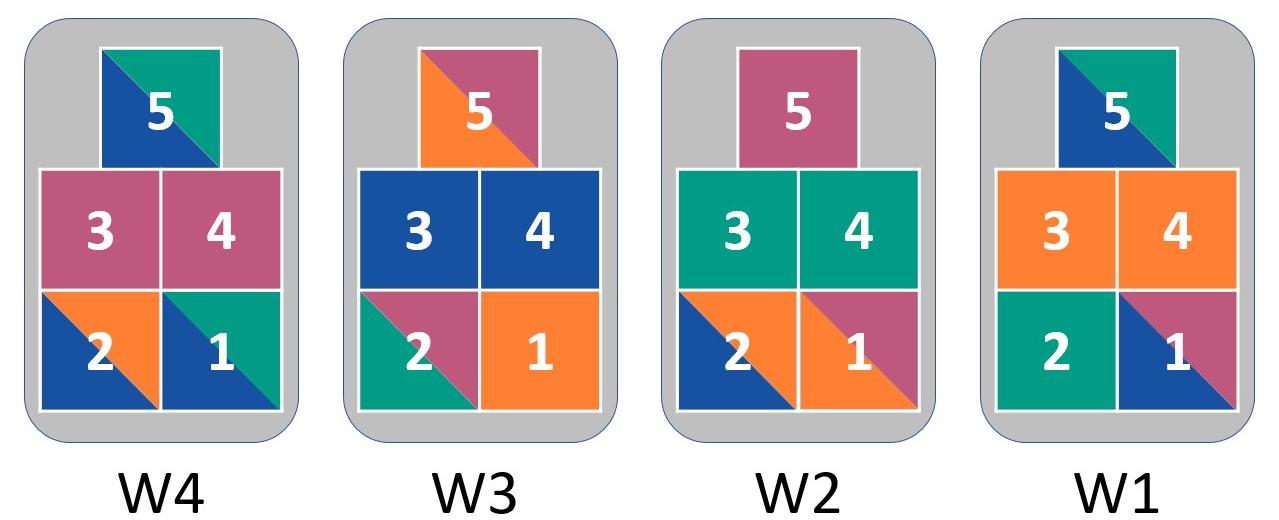}}

  \subfloat[Pattern B\textsubscript{1} ]{%
        \includegraphics[width=0.45\linewidth]{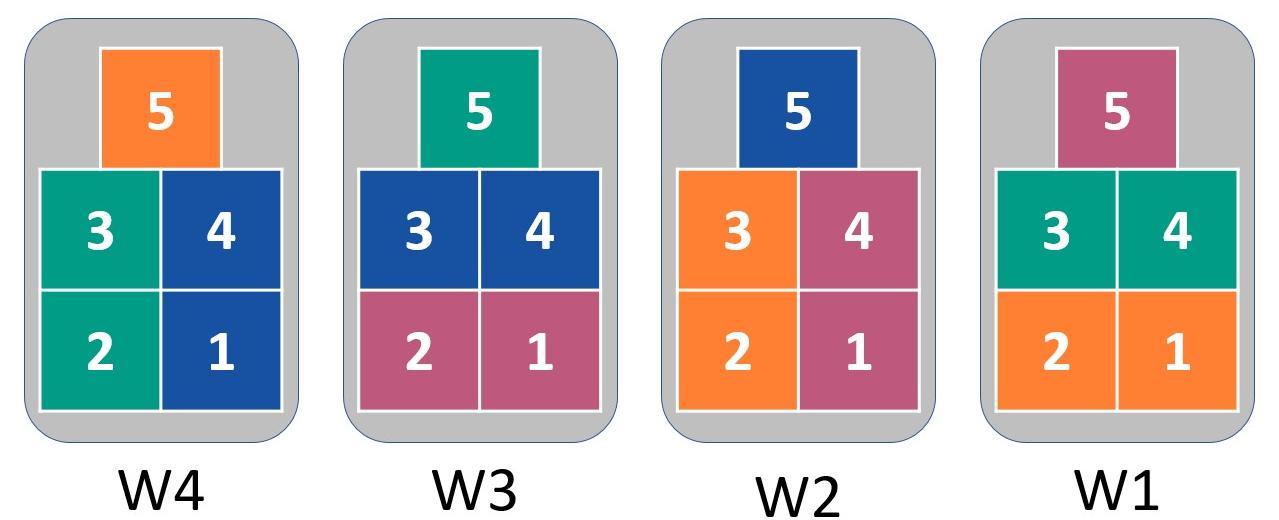}}
    \hfill
  \subfloat[Pattern B\textsubscript{2} ]{%
        \includegraphics[width=0.45\linewidth]{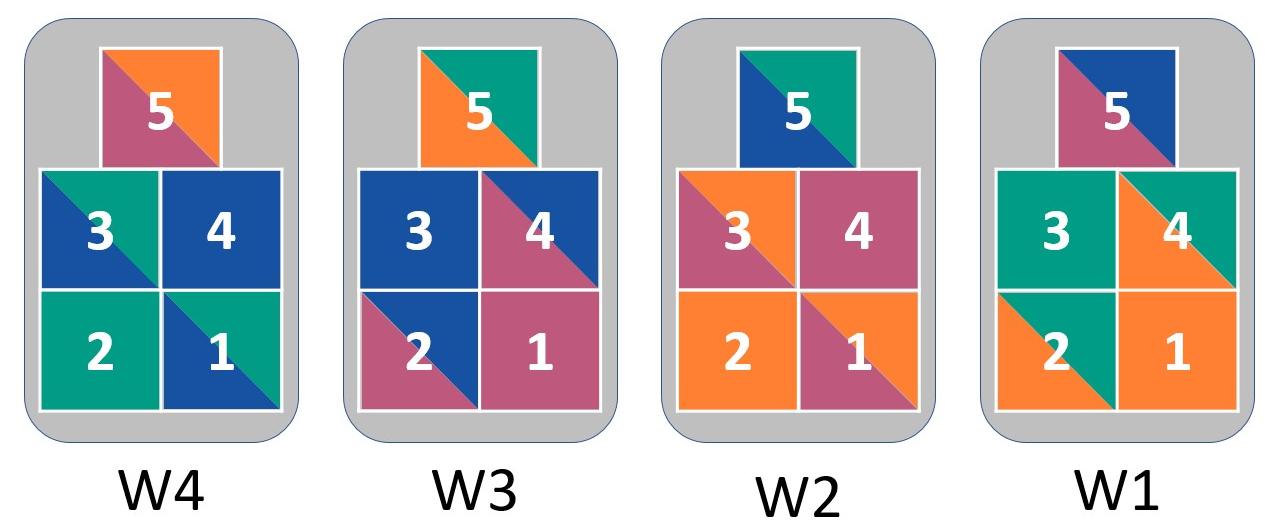}}

  \subfloat[Pattern C\textsubscript{1} ]{%
        \includegraphics[width=0.45\linewidth]{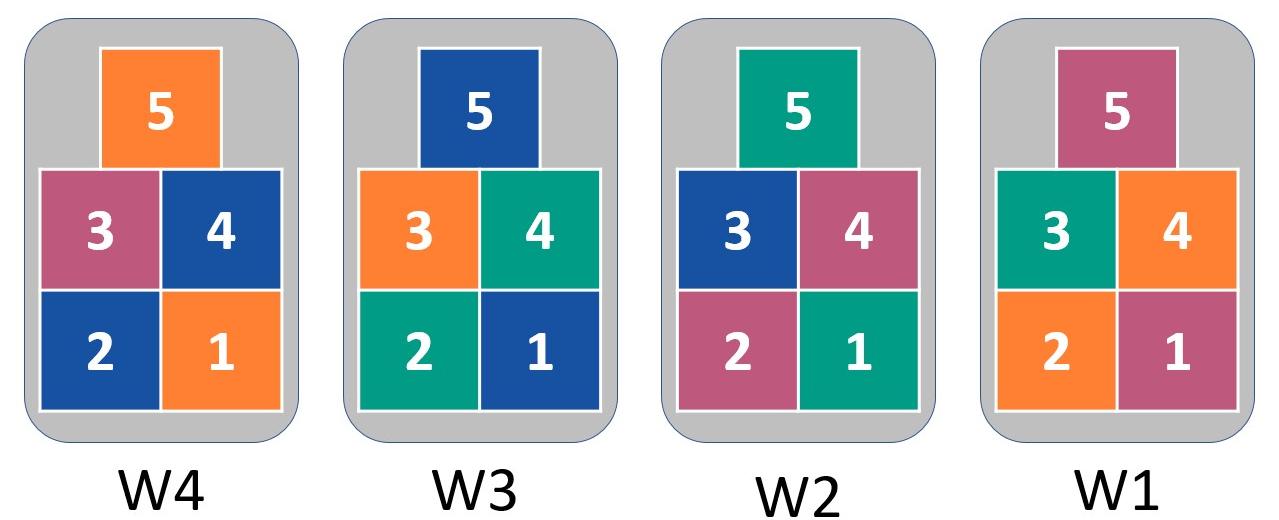}}
    \hfill
  \subfloat[Pattern C\textsubscript{2} ]{%
        \includegraphics[width=0.45\linewidth]{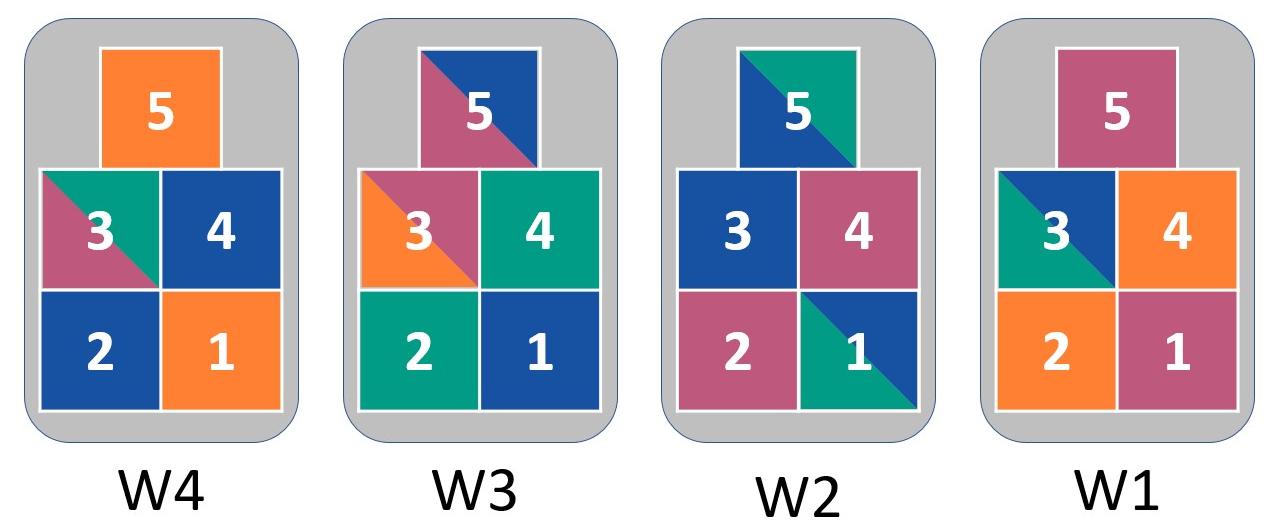}}
        
  \subfloat[Pattern D\textsubscript{1} ]{%
        \includegraphics[width=0.45\linewidth]{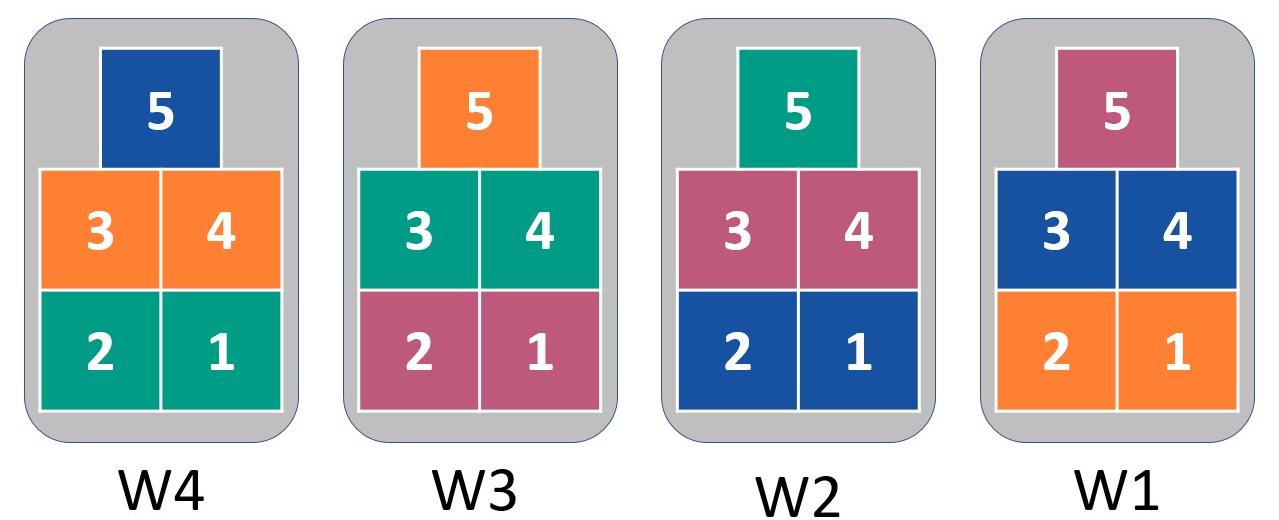}}
    \hfill
  \subfloat[Pattern D\textsubscript{2} ]{%
        \includegraphics[width=0.45\linewidth]{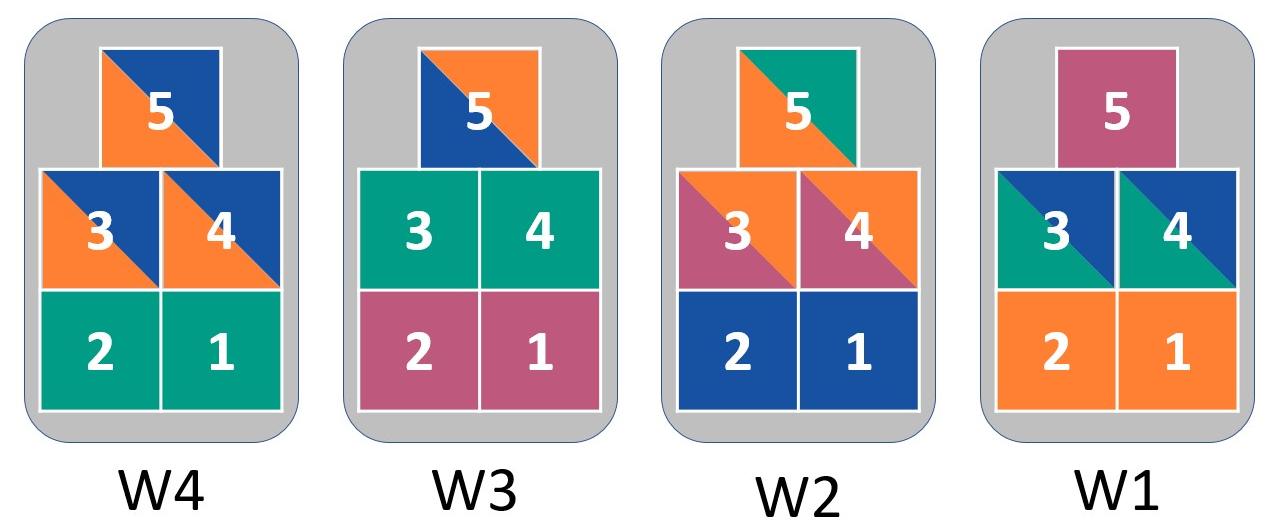}}
  \caption{\textbf{a, c, e, g:} 
Patterns A\textsubscript{1}, B\textsubscript{1}, C\textsubscript{1}, and D\textsubscript{1} represent the set of patterns presented on paper sheets, requiring participants to memorize them within a 45-second timeframe before returning them to the experimenter. \textbf{b, d, f, h:} Patterns A\textsubscript{2}, B\textsubscript{2}, C\textsubscript{2}, and D\textsubscript{2} are variations with some partially known spots, serving as cues for participants throughout the collaborative task, and they are permitted to retain these patterns until task completion to aid in recalling the first pattern.}
  \label{fig:patterns} 
\end{figure}

\begin{itemize}
\item \textbf{Shared area:} The shared area consists of four workspaces denoted as W\textsubscript{1} through W\textsubscript{4}, each containing five numbered spots. As per the predetermined pattern, participants are required to place a colored block in each spot, adhering to the numerical sequence\deleted{,} noted on each spot, from one to five.

\item \textbf{Patterns:} The pattern depicted in Fig.~\ref{fig:pattern_a1} is an example of the pattern the human-robot team must follow when filling the spots. Participants are initially presented with a pattern (Patterns A\textsubscript{1}, B\textsubscript{1}, C\textsubscript{1}, and D\textsubscript{1}) on a sheet of paper, and they are given 45 seconds for memorization. Subsequently, they return the pattern, and the experimenter provides them with a partially known version of the same pattern (Patterns A\textsubscript{2}, B\textsubscript{2}, C\textsubscript{2}, and D\textsubscript{2}), again on a sheet of paper, in which certain spots contain two colors, one of which is correct. This second pattern acts as a cue to recall the initial pattern, and participants are allowed to retain it until the end of the task. Patterns A\textsubscript{2}, B\textsubscript{2}, C\textsubscript{2}, and D\textsubscript{2} feature 9, 12, 6, and 9 partially known spots, respectively, introducing varying levels of difficulty and cognitive load. The decision to incorporate a pattern memorization task was made to introduce a cognitive challenge while adhering to the limitations of a relatively short collaborative scenario. Conducting an extensive experiment that might place excessive mental and physical burdens on participants was considered impractical and challenging to obtain ethics approval. Therefore, concise tasks with durations of around 12-20 minutes were chosen to ensure participants' sustained mental engagement.

\item \textbf{Tables:} The human and robot have separate work areas, with two tables in each. In the human work area, one table is close to the shared area and contains blocks of two colors: green and orange. The other table is far from the shared area (i.e., requires travelling a longer distance) and contains blocks of two other colors: blue and pink. Likewise, the robot's work area has two tables, one distant from the shared area, with blue and orange blocks, and one close to the shared area, including green and pink blocks. Table.~\ref{tab:color_dist} summarizes the distribution of the blocks. Participants were also told to pick up only one block at a time, as the cobot's gripper only has the capacity to pick up a single block. This table and block distribution help evaluate the robot and the human's decision-making and task planning (i.e., selecting and assigning tasks) regarding their preference, travel distance, completion time, and proximity to the optimal plan.

\begin{table}
    \centering
    \caption{Distribution of blocks with respect to the distance to the shared area}
    \label{tab:color_dist}
    \begin{tabular}{@{}l|c|c}
    \toprule
        \textbf{Color} & \textbf{Human} & \textbf{Robot} \\ \midrule\midrule
        Green  & Close & Close \\
        Pink  & Close & Far\\
        Orange & Far & Close \\
        Blue & Far & Far \\
         \bottomrule
    \end{tabular}

\end{table}

\item \textbf{Blocks:} Four unique block colors are employed: green, blue, pink, and orange. The human's tables are equipped with ten blocks of each color, whereas the robot's tables contain eight blocks of each color. This ensures a total of eighteen blocks for each color, exceeding the necessary five blocks per pattern to account for possible errors. A larger quantity of blocks is placed on the human agent's table, taking into account their swifter working pace. Each block has an ArUco marker affixed to it, aiding the robot in locating and handling the blocks within the room.

\item \textbf{Conveyor Belt:} A conveyor belt has been integrated to allow the robot to return any misplaced blocks to the human. The robot places a block at one end of the conveyor belt, which then transports the block to the human's side. This design decision accomplishes two objectives: it separates the workspaces of the human and the robot, mimicking a real industrial environment, while also enhancing safety by reducing the need for the robot and human to interact closely.

\item \textbf{Light Bulb:} On the shared table, there is a red light bulb that lights up when the robot approaches to place or retrieve a block from the shared table. Participants are informed that this light serves as a signal to avoid approaching the table due to safety precautions. However, they can continue planning, navigating within their designated work area, and collecting blocks from other tables. In addition to safety concerns, the light bulb also helps regulate the participants' pace, prompting them to wait for the robot and ensuring that the collaboration is not perceived as a competitive race.

\item \textbf{Safety Measures:} Alongside the safety light bulb, we have set up safety tape and floor cones to create a well-defined physical boundary that separates the work areas of both agents. These measures are complemented by continuous vigilance from the experimenter, who is prepared to intervene as needed. Intervention options include manually controlling the robot using a joystick or promptly activating the emergency safety stop button. During the sessions, there were instances where manual control of the robot was necessary, particularly when it came too close to the tables.

\end{itemize}

\subsubsection{Tasks} Each participant is asked to complete four tasks:
\begin{itemize}
    \item \textbf{Task 0:} In this initial task, participants engage in solo work without the robot's involvement. For all participants, we provide pattern A and follow a consistent procedure. They are tasked with memorizing pattern $\text{A}_1$, printed on a sheet of paper, within a 45-second timeframe and returning it. Subsequently, we furnish them with pattern $\text{A}_2$, which they can retain until the task's conclusion. It is essential to note that errors made by participants in this task do not incur penalties; rather, this task primarily serves to acquaint participants with the setup and offer them practice in block placement. This practice encompasses aspects such as adhering to the correct block order, picking up a single block at a time, and ensuring the markers on the blocks face the robot. While focusing on task practice, we also assess participants' accuracy, self-confidence, and their perception of workload.

    \item \textbf{Tasks 1, 2, 3:} The robot collaborates with the human in these tasks. We consider six distinct permutations (modes) that determine the order of presenting patterns B, C, and D (e.g., {B, C, D}, {C, B, D}). Participants are randomly allocated to one of these six modes (permutations) to ensure an equal distribution among them. Following a consistent procedure, participants are tasked with memorizing the initial pattern within a 45-second timeframe and subsequently returning it. Then, they are provided with the second pattern. To emulate real-world collaborative scenarios in which errors carry consequences, participants are informed that, upon declaring the task as completed, for each block incorrectly placed on the table,  a deduction of \$1 will be applied to their total remuneration. This method of conveying disinformation, classified as a form of deception, received approval from the University of Waterloo Human Research Ethics Board.

\end{itemize}

\subsubsection{Agents' Actions}
We considered a set of six distinct actions for each agent, namely the human and the robot. These actions, outlined in Table~\ref{tab:actions}, are consistent for both agents, ensuring an equivalent level of agency. It is essential to recognize that the feasibility of these actions depends on the current task state, and at each decision point, certain actions may not be applicable. For instance, when the robot has not been assigned any tasks (Action H4), taking on a task delegated by the human is not a viable option. Additionally, to provide the robot with greater autonomy in adjusting its leadership role or reassuming it in the event of suboptimal human performance, we intentionally designed the rejection option (Action H6) to be less straightforward for the human. Action H6 entails the successive execution of H4 and H3 without actually doing them physically. This will be further explained in the next paragraph.

\begin{table*}
    \centering
    \caption{The human and robot's sets of actions}
    \label{tab:actions}
    \begin{tabular}{@{}c|l@{}|c@{}|l@{}}
    \toprule
        \multicolumn{2}{c}{\textbf{Human}} & \multicolumn{2}{c}{\textbf{Robot}} \\ \midrule
        \multicolumn{1}{c}{Action} & \multicolumn{1}{c}{Description} & \multicolumn{1}{c}{Action} & \multicolumn{1}{c}{Description}\\ \midrule
        H1 &Selecting a task for themselves  & R1 &Selecting a task for itself \\
        H2 &Assigning a task to the robot  & R2 &Assigning a task to the human\\
        H3 &Returning \replaced{a}{an} block from the shared workspace & R3 &Returning a \textit{wrong} block from the shared workspace \\
        H4 &Performing a task assigned by the robot & R4 &Performing a \textit{correct} task assigned by the human \\
        H5 & Canceling a task assigned to the robot & R5 &Canceling a task assigned to the human \\
        H6 &Rejecting a task assigned by the robot  & R6 &Rejecting a task assigned by the human\\ \bottomrule
    \end{tabular}

\end{table*}

\subsubsection{Human-robot communication} 
\begin{figure}
    \centering
    \includegraphics[width=\linewidth]{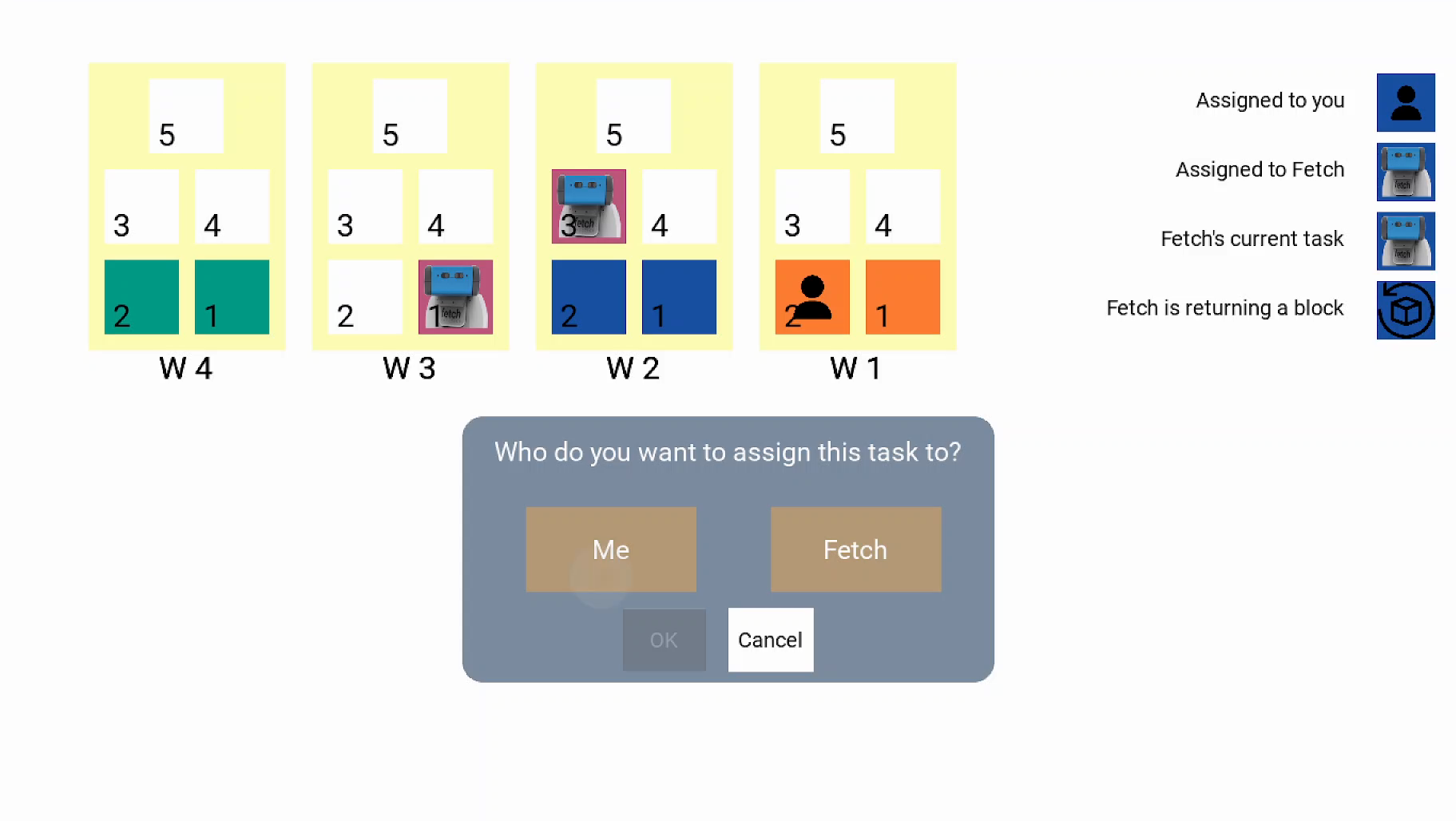}
    \caption{A screenshot of the graphical user interface (GUI) that enables participants to convey their actions to the robot and obtain information regarding the robot's decisions and actions (refer to Table~\ref{tab:actions} for a list of actions).}
    \label{fig:gui}
\end{figure}

Both agents are required to uphold communication to keep each other informed about their forthcoming actions, as specified in Table~\ref{tab:actions}. This communication is facilitated through a purpose-built graphical user interface (GUI) installed on a tablet. Participants have the flexibility to position the tablet on a table within the room or hold it according to their preference. This GUI serves as a medium for participants to assign tasks to the robot and convey information about their intended actions. Similarly, the robot can employ the GUI to assign tasks to the participant and communicate its own planned actions. An image of the GUI is provided in Fig.~\ref{fig:gui}. Notably, the GUI enforces constraints to prevent the human agent from selecting actions that would violate precedence constraints or interfere with tasks already initiated by the robot. Participants received instructions on operating the GUI and had the opportunity to practice using it before commencing the tasks. Additionally, participants were instructed to scan the marker on the blocks before placing them on the shared area, as this action enables the robot to identify the block's ID for potential future retrieval. The GUI streamlines this process by automatically activating the tablet's camera, allowing participants to scan the marker. To maintain brevity, the details of the GUI's design and implementation are omitted.

\begin{remark}
In case participants wish to decline a task assigned by the robot, the sequence of actions involves initially performing the assigned task (H4) and subsequently engaging in the action for returning (H3). Both of these actions are executed through the GUI, without any requirement for physical execution.
\end{remark}

\subsubsection{Robot}
For our experiments, we utilized the Fetch mobile manipulator robot \cite{wise2016fetch}. The robot was configured to perform pick-and-place operations autonomously. While creating a fully autonomous pick-and-place system demanded a considerable amount of time, it was crucial to ensure that participants experienced a work environment resembling real industrial settings. To maintain brevity, we skip the explanation of the autonomous pick-and-place implementation on the robot.

\subsection{Recruitment} We started distributing flyers and recruiting participants after securing ethics approval from the University of Waterloo Human Research Ethics Board. This study involves three phases for each participant.

\subsection{Procedure}
\subsubsection{Phase 0}
The first phase of our study commenced with prospective participants responding to our recruitment flyers and establishing contact with the experimenter. Following this, we sent the participants a consent form via email, along with a questionnaire aimed at gauging their background and experience in the domains of robotics and artificial intelligence.

\subsubsection{Phase 1 (In-Person)}
In Phase 1, we scheduled a 90-minute in-person session with each participant, where we conducted the following procedure:

\begin{enumerate}[label=\textbf{Step \arabic*:}, wide]

  \item We began by welcoming participants and providing an overview of the setup, showing them their designated work area. We used presentation slides to elaborate on the specifics and informed participants that:
  \begin{itemize}
      \item The robot's decision-making may involve occasional errors (deception).
      \item Upon confirming the task's completion, There is a penalty of \$1 for the team per misplaced block  (deception).
  \end{itemize}
  \item \textit{Task 0} - To begin, we gave participants Pattern $\text{A}_1$, as a sheet of paper, and granted them 45 seconds to memorize it. We then collected Pattern $\text{A}_1$ and provided participants with Pattern $\text{A}_2$, a partially known version of Pattern $\text{A}_1$, and indicated that they could keep it until they finished the task. Participants were prompted to assess their self-confidence in successfully completing the task and subsequently initiated the retrieval and placement of the blocks. The experimenter was available to provide guidance and address any deviations from the instructions or questions. It was emphasized that the task had no time constraint, and the completion time would not be evaluated. After successfully placing all the blocks and declaring task completion, the experimenter conducted an inspection of the blocks based on the provided pattern, counting any misplaced blocks. Importantly, participants were assured that no penalties would be applied for errors in this task. Following this, participants were requested to complete a questionnaire regarding their perceived task load.

  \item They watched a video of the pick-and-place of some blocks by the Fetch robot\footnote{\url{https://youtu.be/ahZDo0_iyjg}}, and then answered the questionnaire about their trust in the robot.

  \item Participants familiarized themselves with the GUI and practiced its operation.
  \item \textit{Task 1 } - Depending on the assigned mode (a permutation of patterns B, C, D), participants were provided with the corresponding pattern (e.g., $\text{B}_1$) and instructed to memorize it within a 45-second timeframe. Subsequently, they returned the pattern and received the second pattern, a partially known version of the first pattern (e.g., $\text{B}_2$). Before commencing the task, participants were asked to respond to two questions about their self-confidence and the expected helpfulness of the robot. Following this, they initiated the task, collaborating with the Fetch robot. Upon task completion, participants completed three questionnaires assessing their perceived task load, trust, and perception of the robot.
  \begin{remark}
      Participants initiate the task, and the robot awaits their instructions. They can assign tasks to the robot, and the robot starts working when participants allocate a task to themselves. This approach allows the robot to establish an initial understanding of participants' preferred leading/following roles in the collaboration.
  \end{remark}
  \item \textit{Task 2} - This task followed the same procedure as \textit{Task 1}.
  \item \textit{Task 3} - This task followed the same procedure as \textit{Task 1}.

\item  Lastly, participants were asked to complete two questionnaires. The first questionnaire aimed to evaluate their performance as a team with the robot, while the second one explored their collaborative experience and utilized the shortened version of the User Experience Questionnaire (UEQ)\cite{ueq, ueqshort}. Additionally, participants were asked to rank the difficulty of the tasks (Tasks 0-3) and respond to an open-ended question: \textit{``Which abilities would you improve or add to Fetch if you were to use it in a manufacturing setting?"}.

\end{enumerate}

\subsubsection{Phase 2 (Online)}

In the online phase, we presented each participant with a video recording of their interaction with the robot, with a specific focus on Pattern B. These videos included synchronized content from two different camera angles, a screen recording of the graphical user interface (GUI), and the display of Patterns $\text{B}_1$ and $\text{B}_2$. One of the participant's videos is accessible online\footnote{\url{https://youtu.be/X6Rj0zwQhz8}}. During the online interview, we played the video and encouraged participants to discuss their strategies, plans, and preferences, and how these evolved throughout their collaboration. We also requested them to complete two questionnaires regarding their leadership and followership styles. In line with the approved ethics application, we also informed participants about the existence of \textit{``Deception"} elements in the study and sought their consent to use their data. To express our gratitude for their participation, participants received a \$30 gift card.

\subsection{Measures}

As part of our analysis aimed at assessing participants' perception of the robot, tasks, and collaboration, we utilized a range of subjective measures, primarily derived from the items in the questionnaires they completed at various phases of the study, as well as insights from interviews. Additionally, we considered objective measures such as the number of tasks allocated to the robot by the human.

To assess participants' trust in the robot, we employed Muir's questionnaire \cite{muir1996trust} (refer to Table~\ref{tab:muir}) using a ten-point Likert scale. We administered this questionnaire at multiple points in the study, including after they watched a video of the robot (to gauge initial trust) and following each collaboration with the robot in Tasks 1, 2, and 3.
In addition, participants provided responses to two questions on a 21-point Likert scale, as shown in Table~\ref{tab:selfconf}. These questions pertained to their self-confidence in completing each task (Tasks 0-3) and their expected helpfulness of the robot (Tasks 1-3). Participants answered these questions after reviewing the patterns but prior to commencing the task of fetching and placing the blocks. High and low values on the Likert scale represented high and low self-confidence, as well as the expected helpfulness of the robot.

Following the completion of each task, participants evaluated their workload, considering factors such as perceived mental demand, physical demand, temporal demand, performance, effort, and frustration. They did so by responding to the NASA-TLX questionnaire, as shown in Table~\ref{tab:nasa}, employing a scale ranging from 0 to 100. Additionally, participants completed a questionnaire designed to assess aspects of their working alliance, the traits of the robot teammate, the robot's performance, and the collaboration experience.
The first seven questions in this questionnaire, relevant to our study, were adapted from \cite{hoffman2019evaluating}. This questionnaire, as displayed in Table~\ref{tab:post_percep}, used a 7-point Likert scale, presented in an agree-disagree format.

At the end of the last task (Task 4), participants were asked to complete the user experience questionnaire (UEQ). The UEQ encompasses six paired aspects, each evaluated on a 7-point semantic differential scale: obstructive/supportive, complicated/easy, inefficient/efficient, clear/confusing, boring/exciting, not interesting/interesting, conventional/inventive, and usual/leading edge. Participants also answered a 7-point Likert scale questionnaire about team fluency (i.e., efficient, synchronized collaboration with seamless coordination in shared activities \cite{hoffman2019evaluating}) and their willingness to collaborate again with fetch.

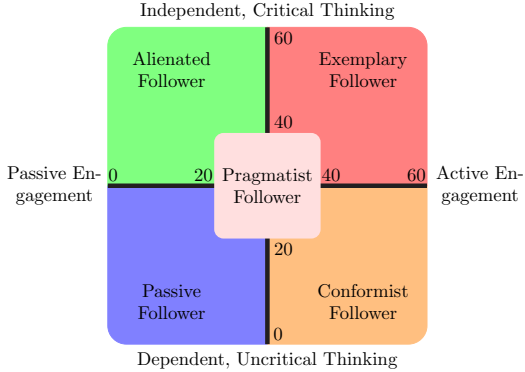
\begin{figure}[!t]
\centering
\scalebox{0.7}{
\begin{tikzpicture}
    \fill [blue!50,draw] (1,0) -- ++(3,0) -- ++(0, -3) {[rounded corners=10] -- ++(-3,0)} -- cycle {};
    \fill [green!50,draw] (1,0) -- ++(3,0) -- ++(0, 3) {[rounded corners=10] -- ++(-3,0)} -- cycle {};
    \fill [red!50,draw] (7,0) -- ++(-3,0) -- ++(0, 3) {[rounded corners=10] -- ++(3,0)} -- cycle {};
    \fill [orange!50,draw] (7,0) -- ++(-3,0) -- ++(0, -3) {[rounded corners=10] -- ++(3,0)} -- cycle {};
    
    \draw [line width=0.85mm, Black] (1, 0) -- (7, 0);
    \draw [line width=0.85mm, Black] (4, -3) -- (4, 3);
    \fill[pink!50, rounded corners=5] (3,-1) rectangle ++(2,2);

    \node[text width=1.8cm, align=center] at (1.1,0.2) {\normalsize \baselineskip=6pt 0};
    \node[text width=1.8cm, align=center] at (2.8,0.2) {\normalsize \baselineskip=6pt 20};
    \node[text width=1.8cm, align=center] at (5.2,0.2) {\normalsize \baselineskip=6pt 40};
    \node[text width=1.8cm, align=center] at (6.8,0.2) {\normalsize \baselineskip=6pt 60};
    \node[text width=1.8cm, align=center] at (4.3,2.8) {\normalsize \baselineskip=6pt 60};
    \node[text width=1.8cm, align=center] at (4.2,-2.8) {\normalsize \baselineskip=6pt 0};
    \node[text width=1.8cm, align=center] at (4.3,-1.2) {\normalsize \baselineskip=6pt 20};
    \node[text width=1.8cm, align=center] at (4.3, 1.2) {\normalsize \baselineskip=6pt 40};
    
    \node[text width=1.8cm, align=center] at (0,0) {\normalsize \baselineskip=6pt Passive Engagement};
    \node[text width=1.8cm, align=center] at (8,0) {\normalsize \baselineskip=6pt Active Engagement};
    \node[text width=5cm, align=center] at (4,3.3) {\normalsize \baselineskip=6pt Independent, Critical Thinking};
    \node[text width=5cm, align=center] at (4,-3.3) {\normalsize \baselineskip=6pt Dependent, Uncritical Thinking};

    \node[text width=1.8cm, align=center] at (2.2,2.2) {\normalsize \baselineskip=6pt Alienated Follower};
    
    \node[text width=1.8cm, align=center] at (5.8,2.2) {\normalsize \baselineskip=6pt Exemplary Follower};

    \node[text width=1.8cm, align=center] at (2.2,-2.2) {\normalsize \baselineskip=6pt Passive Follower};
    
    \node[text width=1.8cm, align=center] at (5.8,-2.2) {\normalsize \baselineskip=6pt Conformist Follower};
    
    \node[text width=1.8cm, align=center] at (4,0) {\normalsize \baselineskip=6pt Pragmatist Follower};

\end{tikzpicture}}
\caption{Kelley's five followership styles}
\label{fig:followership}
\end{figure}

During the online phase of the experiment, we conducted interviews with participants and categorized them into four distinct groups based on their preference for leading or following in collaboration with the robot. These categories included ``lead," ``collaborative-lead," ``collaborative-follow," and ``follow." Additionally, we administered two questionnaires to assess their leadership and followership styles. The leadership style questionnaire utilized a 5-point Likert scale \cite{northouse2014introduction}. Scores for each leadership style (authoritarian, democratic, laissez-faire) reflect the degree of inclination toward that style, with potential scores ranging from very low (6-10) to very high (26-30), as well as intermediate levels such as low (11-15), moderate (16-20), and high (21-25).
To determine participants' followership style, we employed a seven-point Likert scale questionnaire from \cite{kelley1992power}, which determines the score for independent thinking and active engagement. Based on participants' responses, they received scores for independent thinking and active engagement, with scores ranging from 0 to 60. Subsequently, as illustrated in Fig.~\ref{fig:followership}, they were categorized into one of five distinct followership styles.

\begin{table}
    \centering
    \caption{Muir's questionnaire - measuring participants' trust in the robot \cite{muir1996trust}}
    \label{tab:muir}
    \begin{tabularx}{\columnwidth}{@{}p{0.95\columnwidth}@{}}
        \toprule
        1- To what extent can the robot’s behavior be predicted from moment to moment?  (predictability)\\
        2- To what extent can you count on the robot to do its job? (dependability) \\
        3- What degree of confidence do you have that the robot will be able to cope with similar situations in the future? (reliability over time)\\
        4- Overall, how much do you trust the robot? (overall degree of trust)\\
        \bottomrule
    \end{tabularx}

\end{table}

  \begin{table*}
    \centering
    \caption{The questions about the participants' self-confidence and expected helpfulness of the robot. Asked after seeing the patterns but before starting the task}
    \label{tab:selfconf}
    \begin{tabular}{@{}l}
        \toprule
        \textbf{Q1 (Self-confidence):} To what extent are you confident about your ability to successfully accomplish the task?  \\
        \textbf{Q2 (Expected Helpfulness):}  To what extent do you think Fetch can help you improve your performance? \\
        \bottomrule
    \end{tabular}
    
\end{table*}

 \begin{table*}
    \centering
    \caption{NASA-TLX questionnaire asked after each task}
    \label{tab:nasa}
    \begin{tabular}{@{}l}
        \toprule
        \textbf{(Mental demand)} How mentally demanding was the task?  \\
        \textbf{(Physical demand)}  How physically demanding was the task? \\
        \textbf{(Temporal demand)}  How hurried or rushed was the pace of the task? \\
        \textbf{(Performance)}  How successful were you in accomplishing what you were asked to do? \\
        \textbf{(Effort)}  How hard did you have to work to accomplish your level of performance? \\
        \textbf{(Frustration)}  How insecure, discouraged, irritated, stressed, and annoyed were you? \\
        \bottomrule
    \end{tabular}
    
\end{table*}

\begin{table}
    \centering
    \caption{A Questionnaire Assessing Working Alliance, Robot Teammate Traits, Robot's Performance, and Collaboration. Q1-Q7 adopted from \cite{hoffman2019evaluating}}
    \label{tab:post_percep}
    \begin{tabularx}{\columnwidth}{@{}p{0.95\columnwidth}@{}}
        \toprule
        \textbf{Q1:} Fetch was intelligent\\
        \textbf{Q2:} Fetch was committed to the task\\
        \textbf{Q3:} Fetch perceived accurately what my goals are\\
        \textbf{Q4:} Fetch did not understand how I wanted to do the task.\\
        \textbf{Q5:} Fetch and I worked towards mutually agreed-upon goals.\\
        \textbf{Q6:} Fetch and I respected each other.\\
        \textbf{Q7:} I feel that Fetch appreciates me.\\
        \textbf{Q8:} Fetch and I collaborated well together.\\
        \textbf{Q9:} Fetch worsened the team's performance in terms of completion time.\\
        \textbf{Q10:} Fetch improved the team's performance in terms of accuracy.\\
        \textbf{Q11:} The robot's decisions were reliable.\\
        \textbf{Q12:} The robot failed to pick up and place objects reliably.\\
        \bottomrule
    \end{tabularx}

\end{table}

 \begin{table}
    \centering
    \caption{The robot and team's overall performance. Q1-3 are adopted from \cite{hoffman2019evaluating}}
    \label{tab:fluency}
    \begin{tabularx}{\columnwidth}{@{}p{0.95\columnwidth}@{}}
        \toprule
        1- Our team improved over time.  \\
        2- Our team's fluency improved over time.\\
        3- The robot's performance improved over time.\\
        4- I will be happy to collaborate again with this robot in similar future tasks\\
        \bottomrule
    \end{tabularx}

\end{table}

\section{Results \& Discussion}
We recruited 58 participants from our university. However, we had to omit data from 10 participants due to reasons such as a bug in the robot's program and the robot's malfunction. Consequently, our data analysis is centered on the remaining 48 participants, comprising 22 females, 24 males, and 2 individuals self-identifying as 'others,' with an average age of 24.02~$\pm$~3.93. The majority were University students (44), while 3 were postdoctoral or visiting researchers, and 1 was a staff member. 
Initially, we delve into participants' perception of the tasks, the robot, and collaboration. Subsequently, we shift our focus to the analysis of participants' actions and performance. The discussion pertaining to the robot's actions and performance is deferred to another paper centering more explicitly on the planning framework \cite{fetch_robot}.

\subsection{Participants' Perception of the Tasks, Robot, \& Collaboration}
This part mainly focuses on subjective measurements through questionnaires. Throughout this section, the Kruskal–Wallis H test, a nonparametric statistical test, is applied to ascertain if there are statistically significant variations among two or more groups. If a noteworthy overall difference is detected among multiple groups, the Dunn test is subsequently employed as a post hoc analysis to pinpoint specific group differences. In addition, whenever we state that the analysis is based on the tasks, it implies an examination following the chronological sequence, starting from Task 0 and progressing through Task 3.

\subsubsection{Trust}
\begin{figure}
    \centering
    \includegraphics[width=0.8\linewidth]{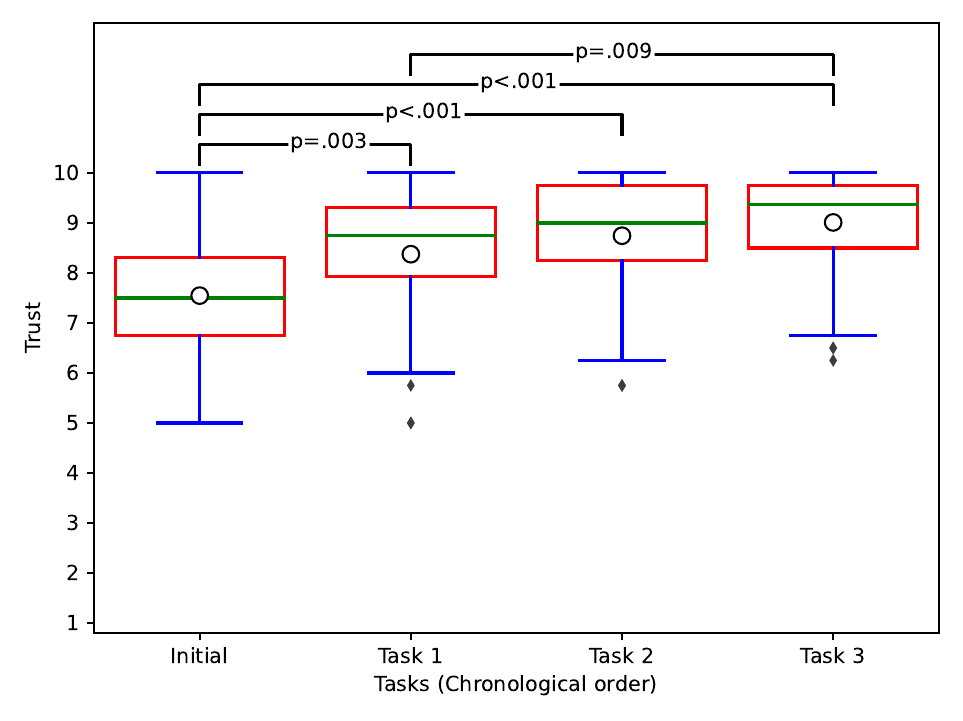}
    \caption{Participants' trust in the robot before collaboration and after each task}
    \label{fig:result_trust}
\end{figure}

We assessed participants' trust in the robot at four stages: before collaborating with the robot, after watching a video of the robot (initial trust), and after each collaboration for Tasks 1, 2, and 3. We found a statistically significant difference among these stages ($H(3) = 13.85$, $p = .003$). Fig.~\ref{fig:result_trust} illustrates participants' trust in the robot, representing the average of their responses to questions in Table~\ref{tab:muir}. The figure reveals a notable difference between participants' initial trust and trust levels after Tasks 1, 2, and 3. Furthermore, it demonstrates a gradual increase in participants' trust over the sessions, with a significant difference observed between Task 1 and Task 3.

\subsubsection{Self-confidence} 
\begin{figure}
    \centering
    \includegraphics[width=0.8\linewidth]{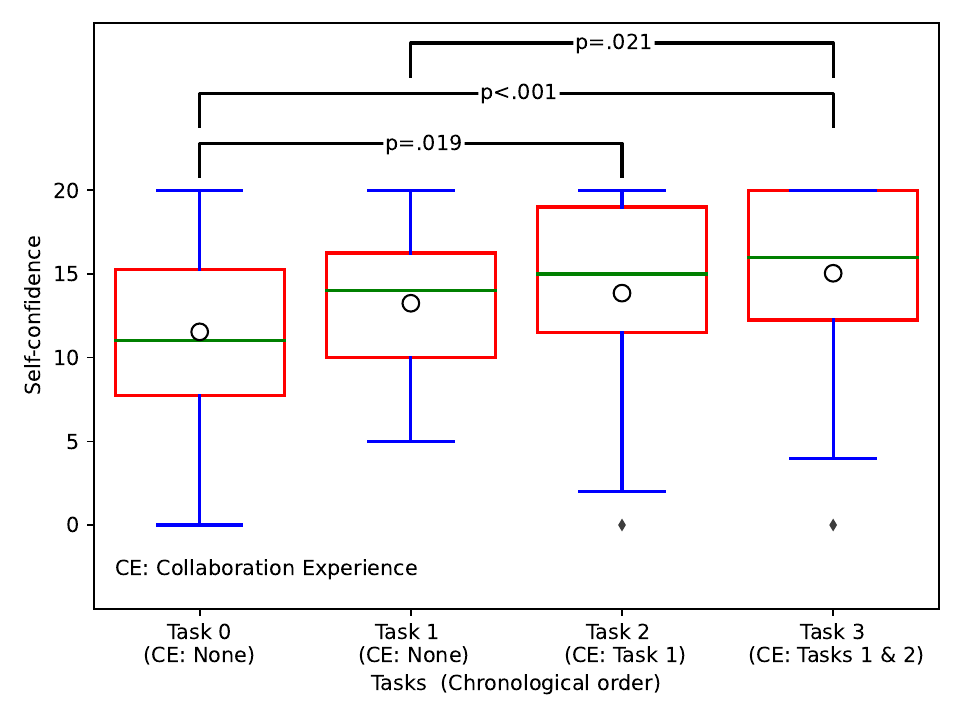}
    \caption{Participants' self-confidence in completing the tasks after seeing the patterns but before starting them based on the tasks}
    \label{fig:result-confidence_task}
\end{figure}

Participants were also asked about their self-confidence in accomplishing the tasks before collaborating with the robot, but after seeing the pattern they needed to work on. In Task 0, participants were told they had to work alone, without the robot. To answer this question in Task 1, participants knew that the robot would work alongside them, but they had not yet worked with the robot, and it was their first experience. Fig.~\ref{fig:result-confidence_task} shows the results for this question, revealing a significant difference in self-confidence across different tasks ($H(3) = 13.85$, $p=.003$). As expected, despite an increase in the average self-confidence of participants in Task 0 and Task 1, there is no significant difference between them, as both tasks involved participants working without prior experience with the Fetch robot. However, the results indicate a significant difference between Task 0 and Task 2, Task 0 and Task 3, as well as Task 1 and Task 3.

\subsubsection{Helpfulness} 
\begin{figure}
        \centering
        \includegraphics[width=0.8\linewidth]{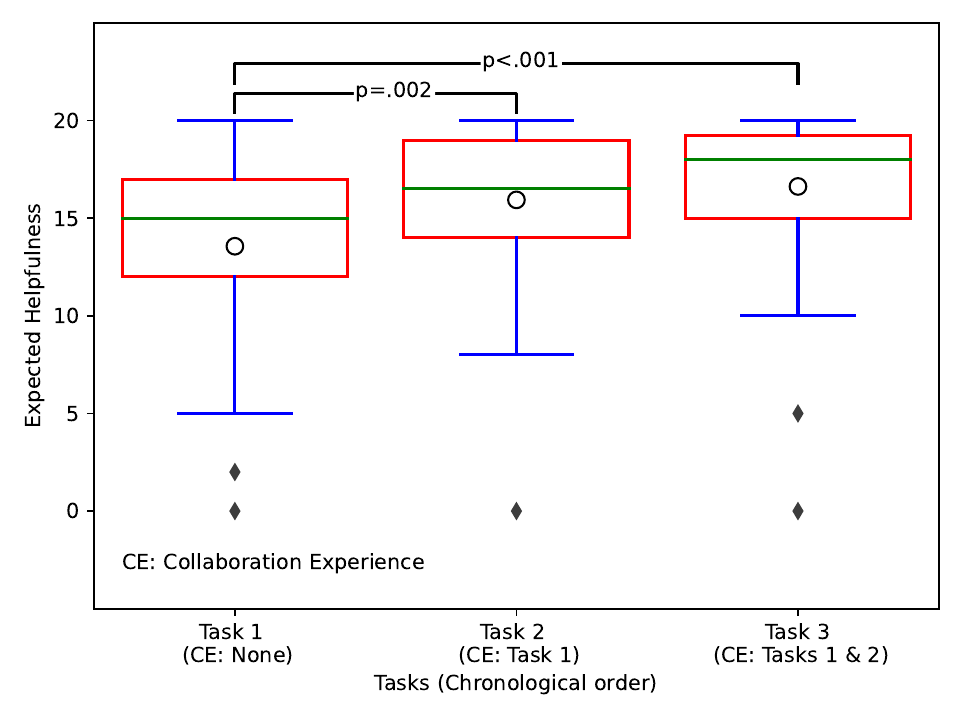}
        \caption{Participants' expected helpfulness from the robot}
        \label{fig:result_helpfulness}
\end{figure}

Participants' self-confidence improvement can be attributed to becoming more familiar with the task and finding the robot more helpful over time. However, to demonstrate that this improvement is not solely due to the former, in Tasks 1, 2, and 3, in addition to inquiring about their self-confidence, we also asked about their expectations regarding the robot's helpfulness in completing the task. This revealed a significant difference among these responses ($H(2) = 19.76$, $p\ll.001$). As depicted in Fig.~\ref{fig:result_helpfulness}, participants' expected helpfulness of the robot increased over time, with a significant difference observed between their expectations in Tasks 1 and 2, as well as Tasks 2 and 3. It is worth noting that when participants answered this question in Task 1, they had not yet collaborated with the robot. Thus, participants perceived the robot as more helpful after the initial collaboration. However, there was no significant difference in participants' expected helpfulness of the robot across different patterns.

\subsubsection{Relative Trust} We also evaluated participants' relative trust, i.e., the difference between participants' trust in the robot and their self-confidence. The result shows no significant difference in participants' relative trust between Tasks 1, 2, and 3.

\subsubsection{Robot's Performance and Traits}
\begin{figure*}
    \centering
    \includegraphics[width=0.9\linewidth]{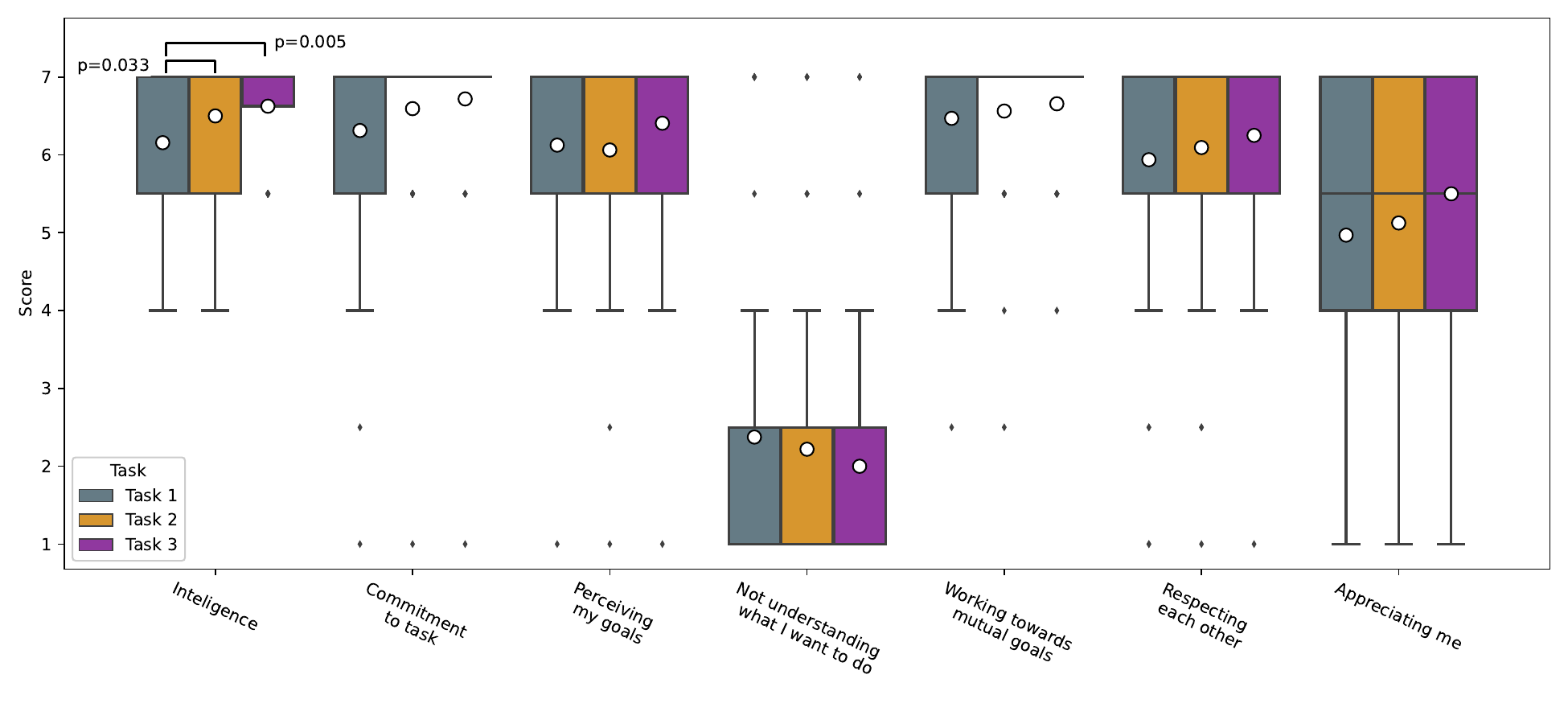}
    \caption{Participants' perception of the robot's traits and working alliance}
    \label{fig:result_tratis}
\end{figure*}

\begin{figure}
    \centering
    \includegraphics[width=\linewidth]{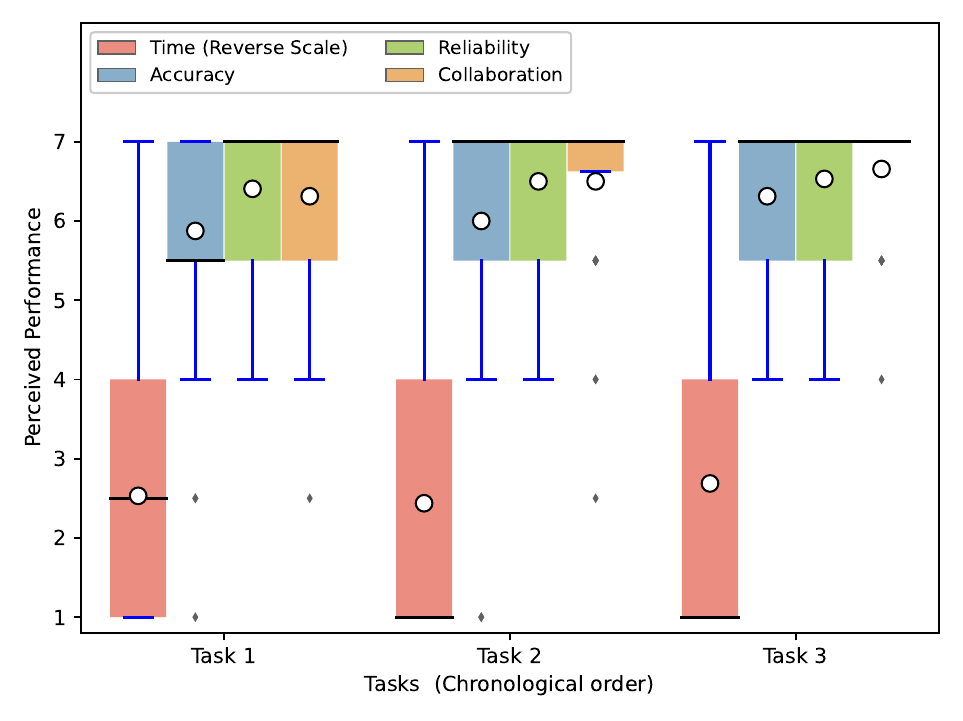}
    \caption{Participants' perception of the robot's performance and collaboration}
    \label{fig:result_trac}
\end{figure}

After each task (Tasks 1-3), participants answered the questions in Table \ref{tab:post_percep}. In Fig.~\ref{fig:result_tratis}, participants' scores indicate a positive working alliance and favorable robot teammate traits. We also observe a significant improvement in how participants perceive the robot's intelligence between Tasks 1 and 2, as well as between Tasks 1 and 3. We asked participants to rate their collaboration with the robot and its performance in terms of reliability, accuracy, and operating speed. Based on Fig.~\ref{fig:result_trac}, participants found the robot's decisions accurate and reliable, leading to a positive collaboration experience. It is worth noting that Fetch was noticeably slower than the human agent in pick-and-place tasks, which is reflected in some participants' scores. However, overall, the average of participants' scores is low (reverse scale), indicating they believed Fetch did not significantly extend task completion time. This perception aligns with participants' understanding of the robot's role in the collaboration, as one participant expressed during the interview: ``Mentally, it can compensate for me, but physically, I can compensate for it (Fetch)."

\subsubsection{User Experience \& Team Fluency}
\begin{figure}
    \centering
    \includegraphics[width=\linewidth]{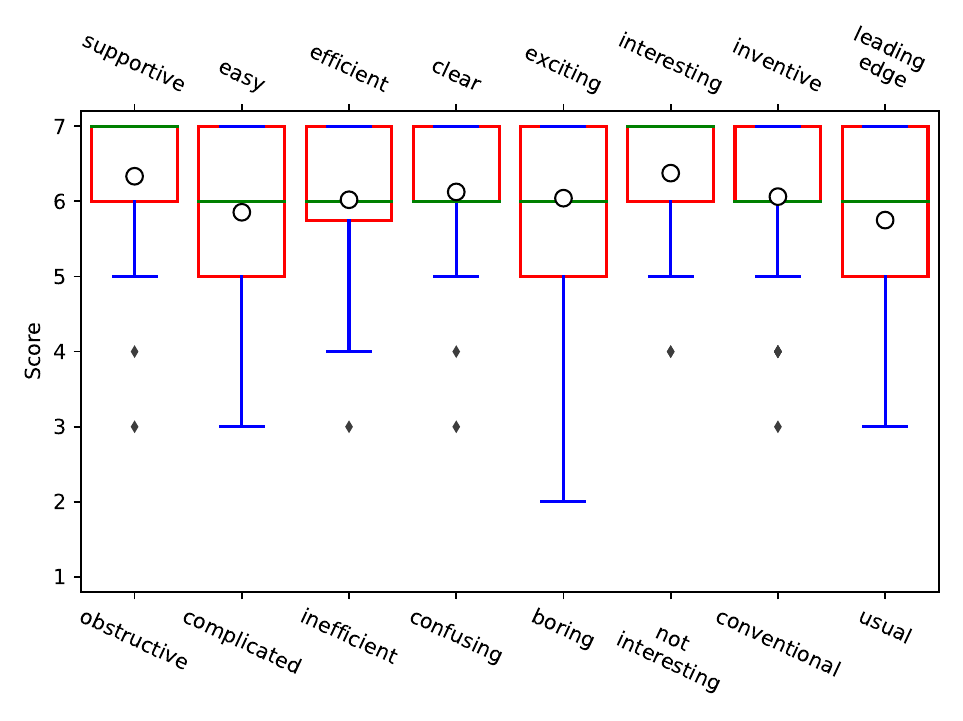}
    \caption{Scores of user experience questionnaire}
    \label{fig:result_ueq}
\end{figure}

\begin{figure}
    \centering
    \includegraphics[width= \linewidth]{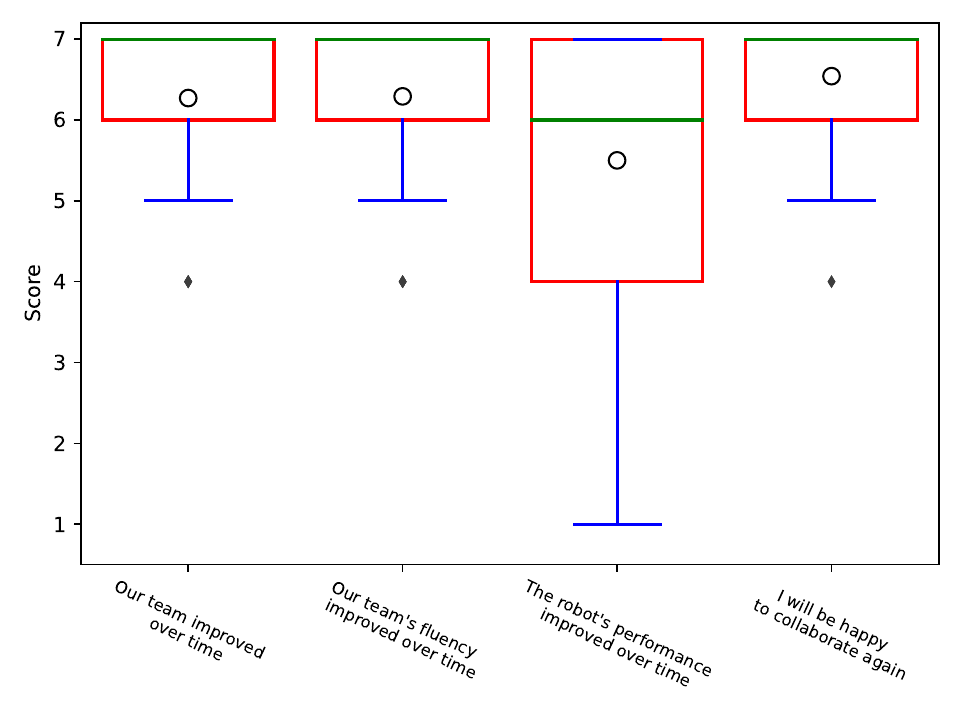}
    \caption{Team fluency and performance}
    \label{fig:result_ovp}
\end{figure}

At the end of the in-person phase of the experiment, participants answered the user-experience questionnaire. Fig.~\ref{fig:result_ueq} shows participants' scores on this questionnaire, indicating a positive user experience. They also answered four questions in Table~\ref{tab:fluency}, and Fig.~\ref{fig:result_ovp} shows their scores for these four questions. Based on it, participants perceived an improvement in their team performance and fluency and expressed a willingness to collaborate again with Fetch. Despite a high average, participants' scores regarding the improvement in the robot's performance cover the whole range of the scale. However, this is reasonable as the robot had the same plan for all three tasks.



\subsubsection{Perceived Workload}
\begin{figure*} 
    \centering
  \subfloat[Mental demand  \label{fig:result_mental_task}]{%
   \includegraphics[width=0.33\linewidth]{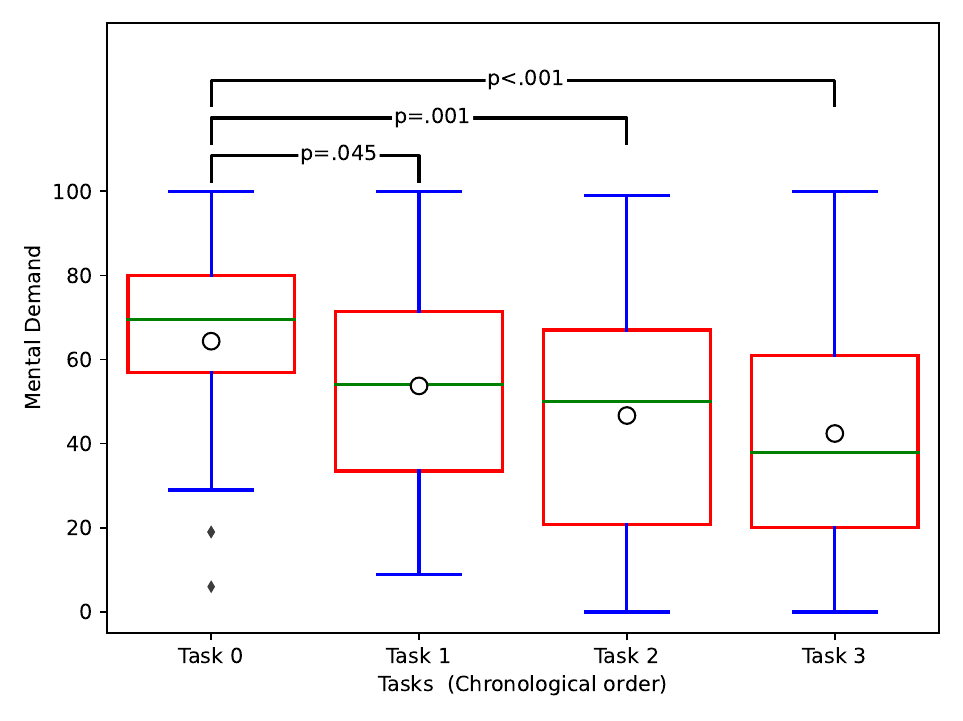}}
  \subfloat[Physical Demand \label{fig:result_physical}]{%
       \includegraphics[width=0.33\linewidth]{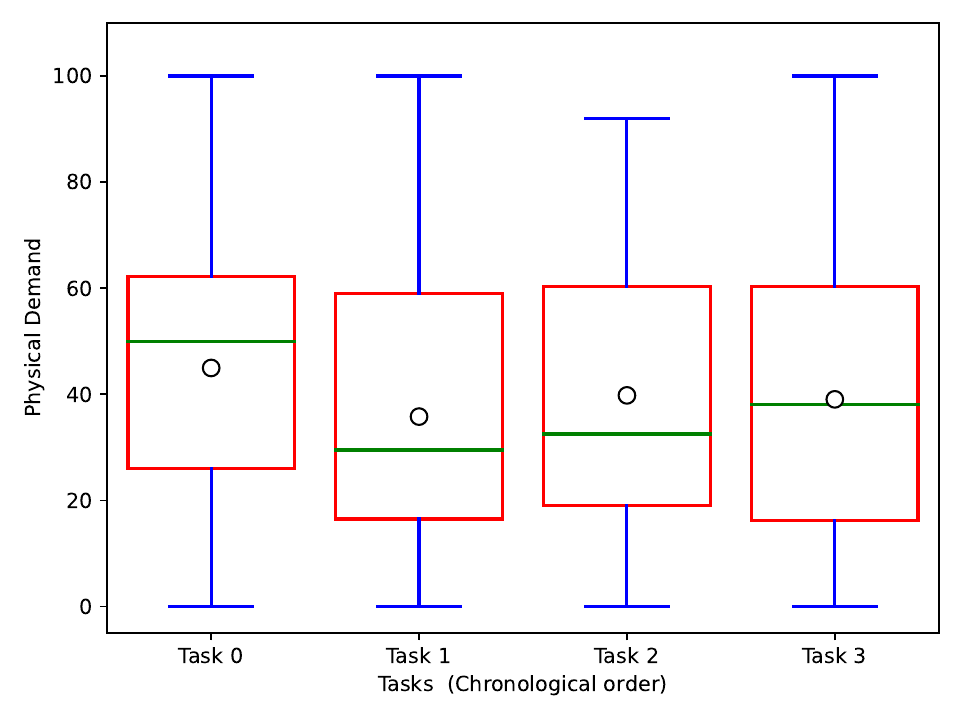}}
  \subfloat[Temporal demand\label{fig:result_temporal}]{%
        \includegraphics[width=0.33\linewidth]{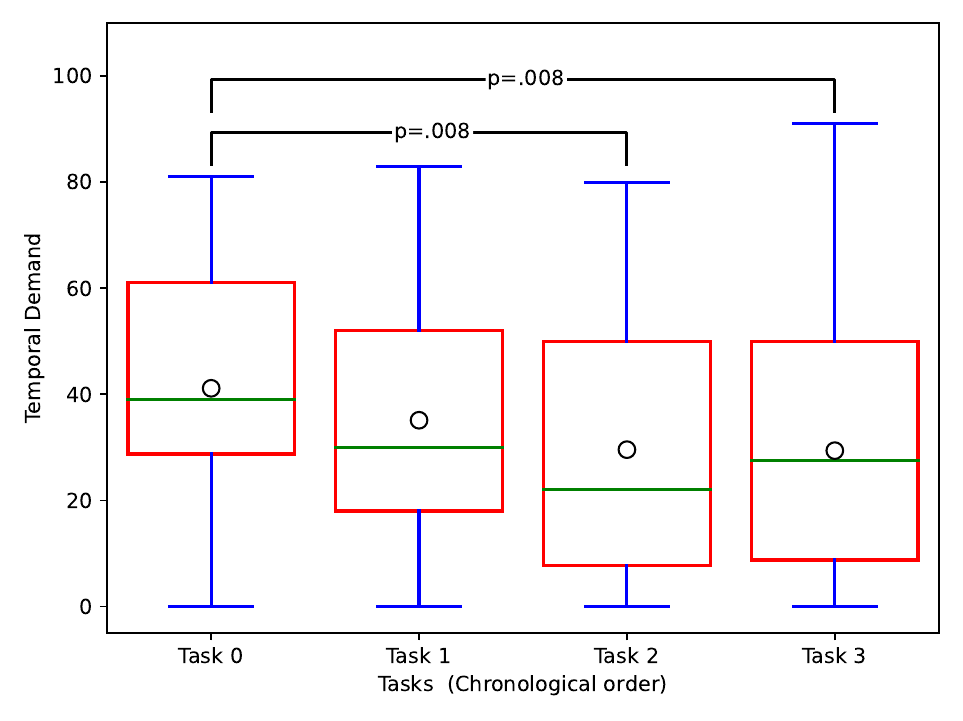}}
        \hfill
  \subfloat[Performance\label{fig:result_nasa_performance}]{%
        \includegraphics[width=0.33\linewidth]{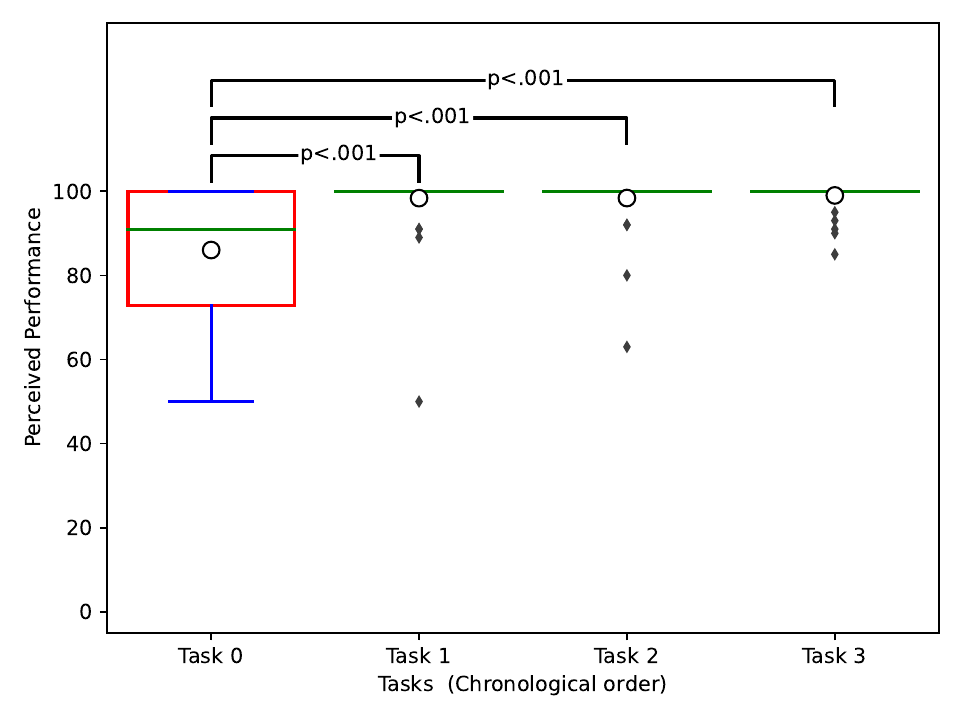}}
  \subfloat[Frustration\label{fig:result_frust}]{%
        \includegraphics[width=0.33\linewidth]{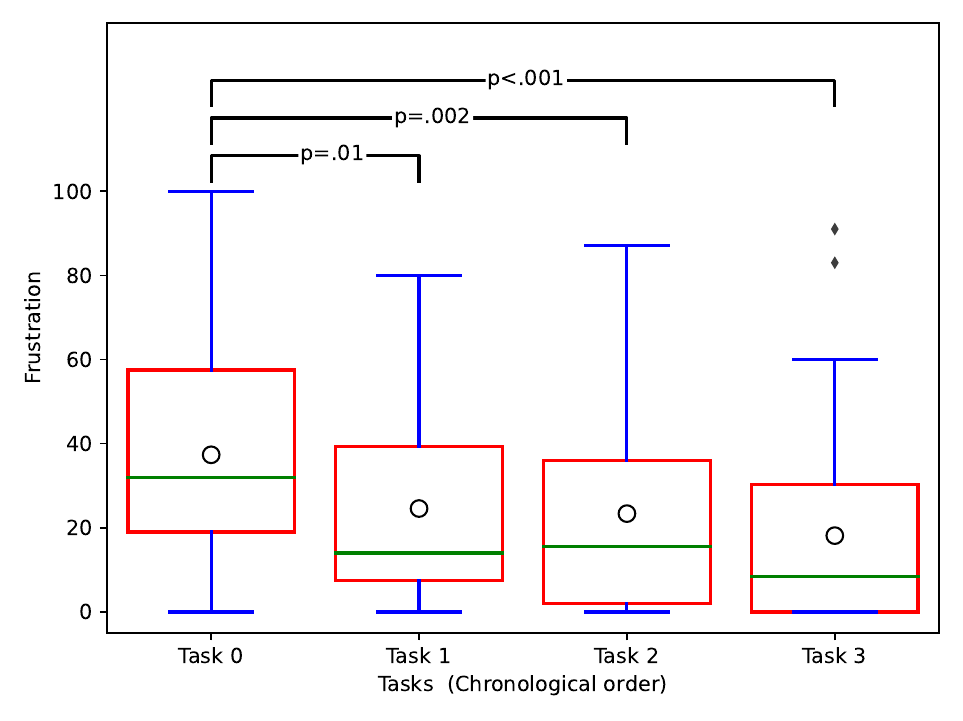}}
  \subfloat[\added{Effort} \label{fig:result_effort_task}]{%
       \includegraphics[width=0.33\linewidth]{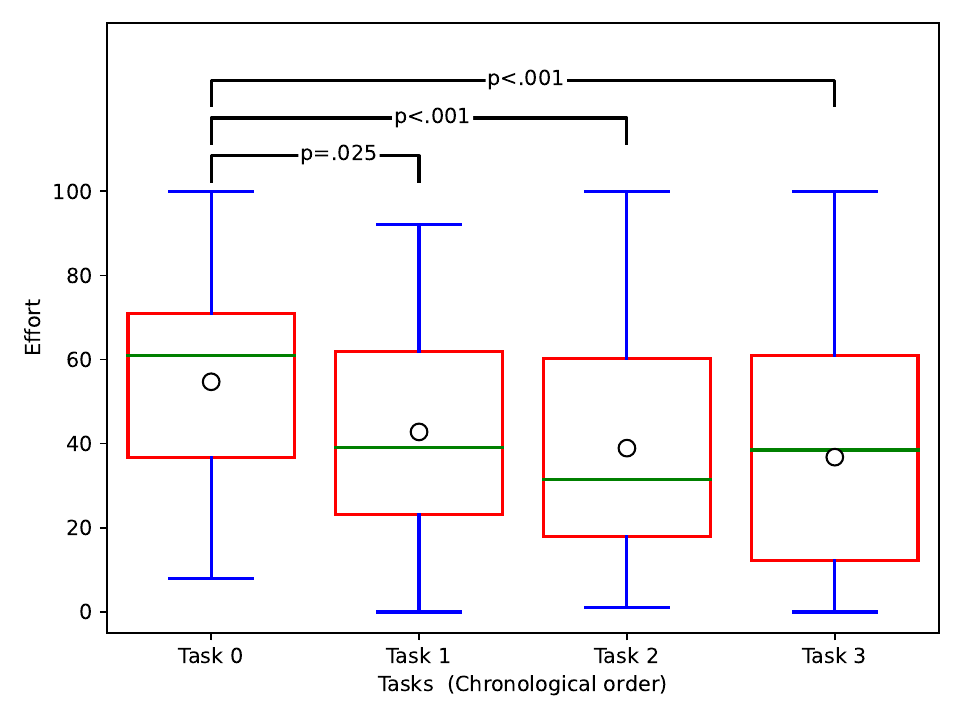}}
  \caption{Participants' perceived a) mental, b) physical demand, c) temporal demand, d) performance,  e) frustration, and f) effort based on the tasks}
  \label{fig:result-nasa4} 
\end{figure*}

After each task (Tasks 0-3), participants answered the NASA-TLX questionnaire. Hereinafter, we discuss the results for each of the six measures in this questionnaire.
\begin{itemize}[leftmargin=*]
    \item \textbf{Mental Demand:} 
The mental demand for each task started from the point when participants were given the first pattern (i.e., Patterns $\text{A}_1$, $\text{B}_1$, $\text{C}_1$, and $\text{D}_1$) to memorize until the end of the task where they needed to recall the pattern with the help of the second given pattern (i.e., Patterns $\text{A}_2$, $\text{B}_2$, $\text{C}_2$, and $\text{D}_2$). We analyzed participants' scores for this question based on the tasks and the patterns. Fig.~\ref{fig:result_mental_task} shows the result for Task 0 to Task 3 (in chronological order), with a significant difference between them ($H(3) = 18.05$, $p<.001$). The results show a significant required mental demand for Task 0 (working alone) compared to Tasks 1, 2, and 3. This indicates that collaboration with the robot could decrease participants' perceived mental demands, supporting part of Hypothesis \ref{hypo4}.

\item \textbf{Physical Demand:}
Fig.~\ref{fig:result_physical} illustrates the results for physical demand scores based on the chronological order of the tasks. Despite the higher mean value of physical demand for Task 0, and in contrast to Hypothesis~\ref{hypo4}, we observe no significant difference between Task 0 and Tasks 1, 2, and 3 ($H(3) = 3.35$, $p=.34$). The reason can be that participants placed most of the blocks on the table, as the robot was slower than them and, therefore, did not experience a higher physical demand compared to Task 0.

\item \textbf{Temporal Demand:} Participants were told several times during the experiment that there was no time limit, and we did not measure their time. However, as shown in Fig.~\ref{fig:result_temporal}, there is a significant difference between them ($H(3) = 9.58$, $p=.023$), and participants perceived a higher temporal demand in Task 0 compared to Tasks 2 and 3, rejecting part of Hypothesis~\ref{hypo4}.

\item\textbf{Performance:} In Tasks 1, 2, and 3, as the robot could fix participants' errors, all participants completed the tasks successfully. Conversely, in Task 0, participants made some mistakes and had no teammate to fix the errors. Therefore, as shown in Fig.~\ref{fig:result_nasa_performance}, participants significantly perceived themselves as more successful in completing Tasks 1, 2, and 3 compared to Task 0 ($H(3) = 52.75$, $p\ll .001$). This supports part of Hypothesis~\ref{hypo4}.

\item \textbf{Frustration:} The results show that there is a significant difference in perceived frustration in the tasks ($H(3) = 19.25$, $p< .001$), specifically in Task 0 in comparison with Tasks 1, 2, and 3 (Fig.~\ref{fig:result_frust}. However, there is no significant difference in perceived frustration between Tasks 1-3 and Patterns B, C, and D ($H(3) = 0.27$, $p=.87$). This result also confirms part of Hypothesis \ref{hypo4}, which mentions that participants experienced less frustration when collaborating with the robot.

\item \textbf{Effort:} The results show a significant difference in perceived effort based on the tasks ($H(3) = 15.74$, $p=.001$) and patterns ($H(3) = 6.89$, $p=.03$). We can see in Fig.~\ref{fig:result_effort_task} that participants perceived more required effort to accomplish Task 0 in comparison with Tasks 1, 2, and 3. This confirms part of Hypothesis 6, which states collaboration can decrease perceived effort. 
\end{itemize}

\subsubsection{Participants' Initial Trust vs. Perception of the Robot}
\begin{figure*}
    \centering
    \includegraphics[width=\linewidth]{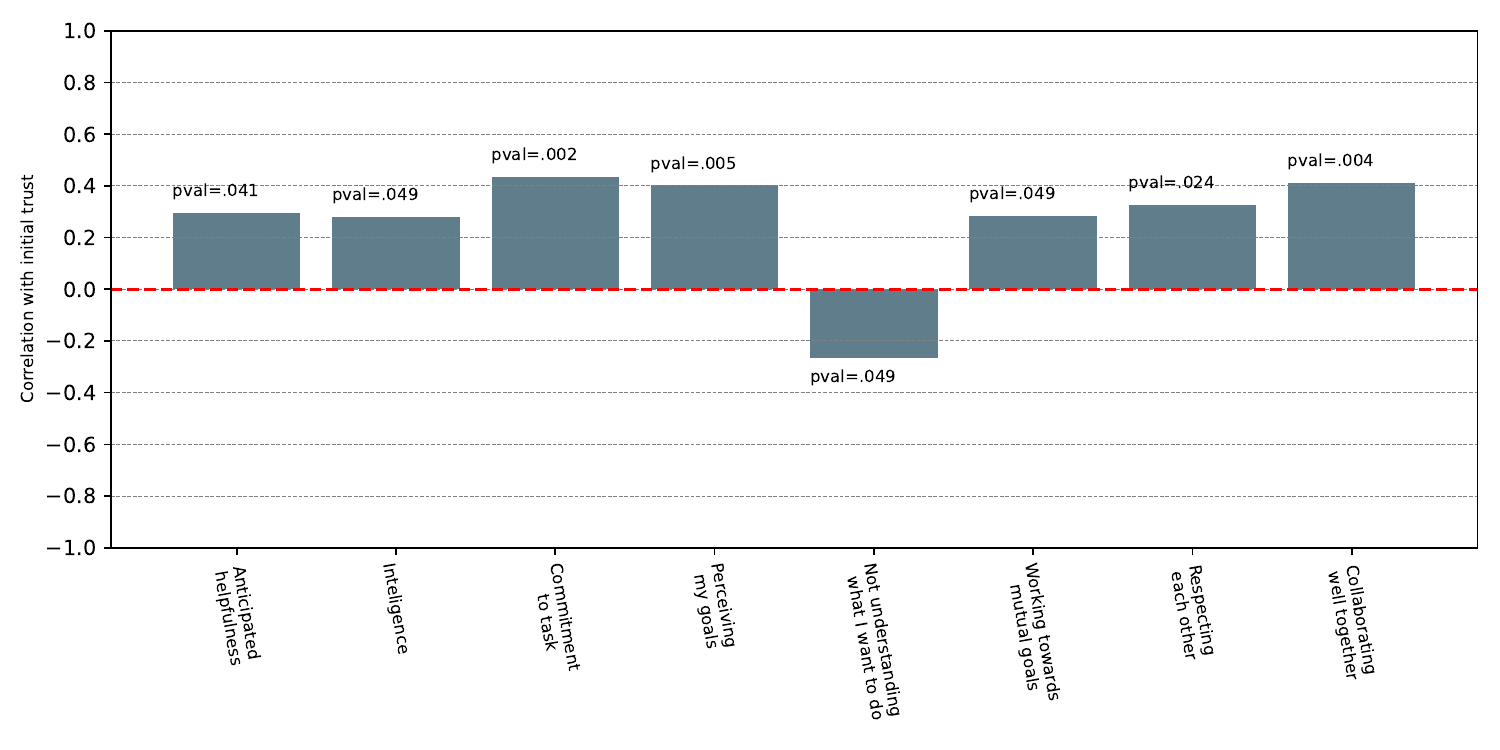}
    \caption{Correlation between participants' initial trust and perceived robot traits}
    \label{fig:result_init_trust}
\end{figure*}

We assessed the correlation between participants' initial trust and their perceptions of the robot's traits, helpfulness, and collaboration (Fig.~\ref{fig:result_init_trust}). The findings indicate weak correlations between participants' initial trust and their expected helpfulness of the robot, perceived robot's intelligence, understanding, and respect for them. Additionally, we observed moderate correlations between their initial trust and perceived robot's commitment to the task, understanding of goals, and collaboration. 

\subsubsection{Task Difficulty}\label{sec:difficulty}
\begin{figure*} 
    \centering
    \subfloat[Mental demand\label{fig:result_mental_pattern}]{
        \includegraphics[width=0.24\linewidth]{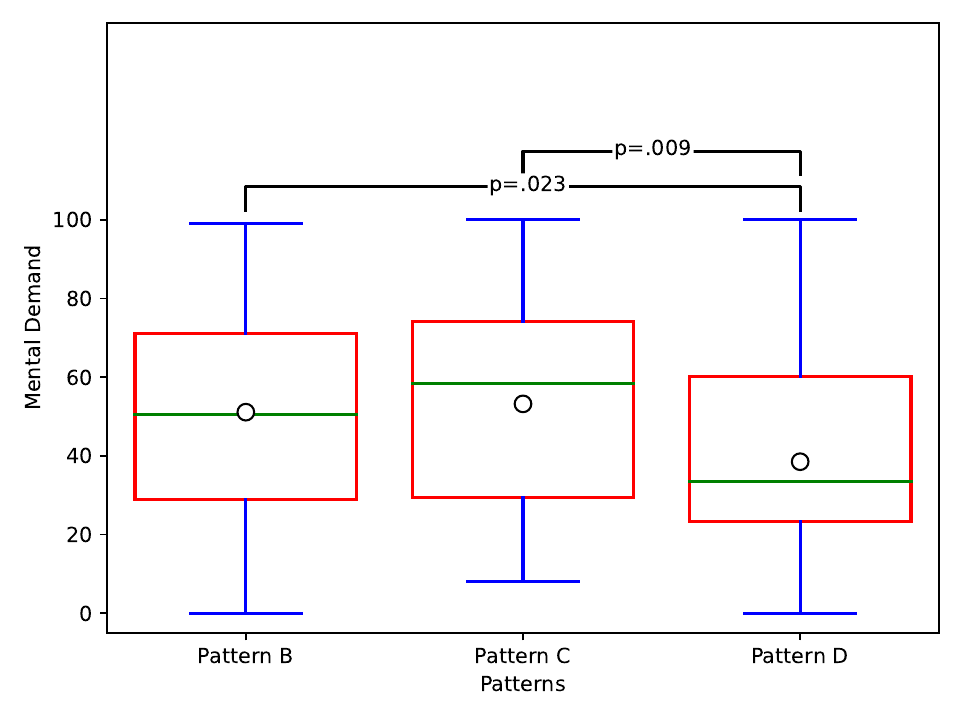}}
    \hfill
    \subfloat[Effort\label{fig:result_effort_pattern}]{
        \includegraphics[width=0.24\linewidth]{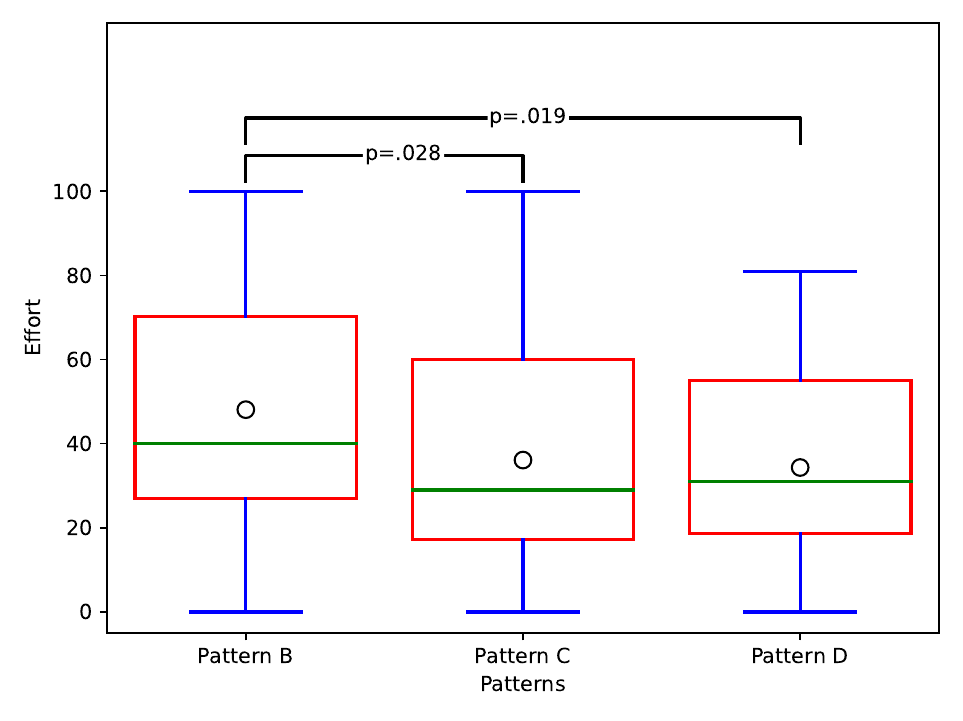}}
    \hfill
    \subfloat[Self-confidence\label{fig:result-confidence_pattern}]{
        \includegraphics[width=0.24\linewidth]{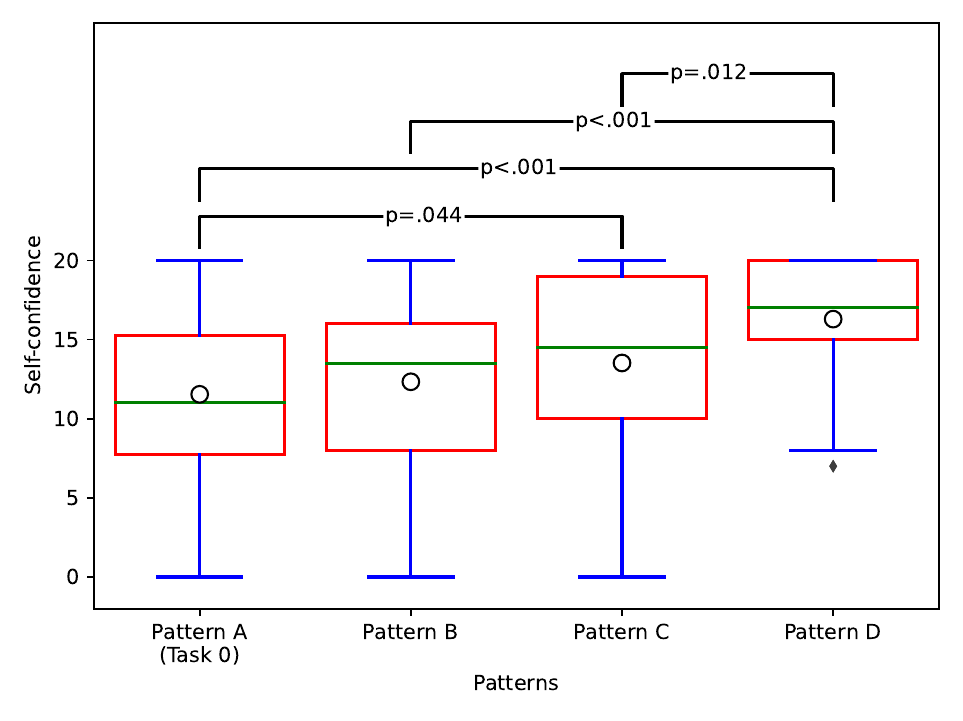}}
    \hfill
    \subfloat[Relative trust \label{fig:result_relative_trust}]{
        \includegraphics[width=0.24\linewidth]{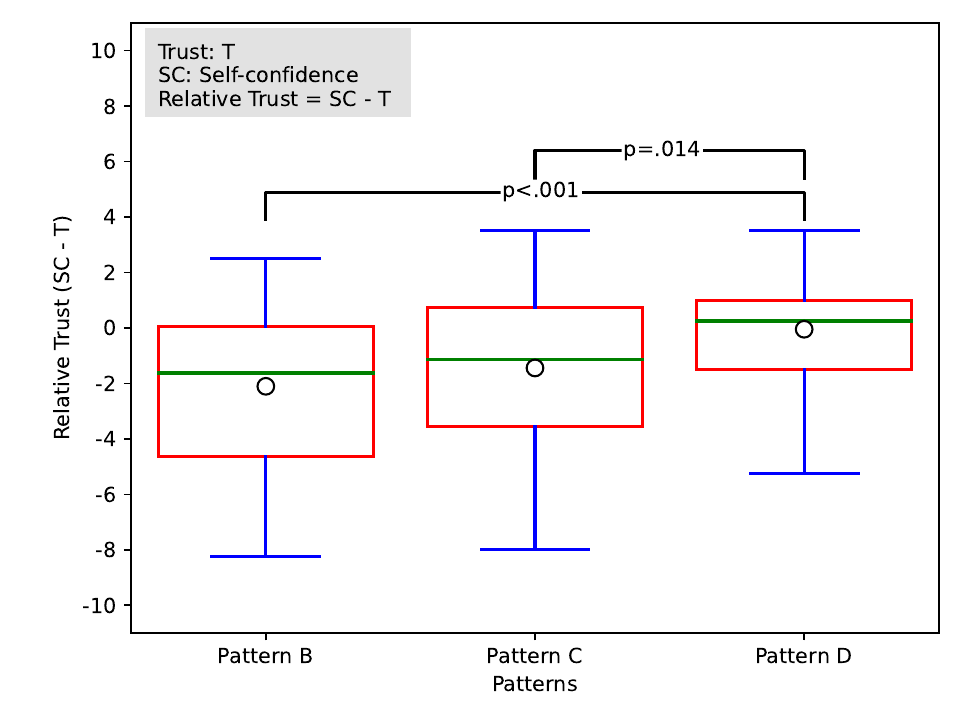}}
    \caption{Participants' a and b) perceived mental demand and effort, c) self-confidence, and d) relative trust (self-confidence - trust in the robot) based on the patterns (Patterns B, C, and D)}
    \label{fig:result-patterns_perception} 
\end{figure*}

Here, we investigate how task difficulty influenced participants' perception of collaboration, performance, and decision-making.
\begin{itemize}[leftmargin=*]
    \item \textbf{Participants Ranking:}
At the end of the last task, we asked participants to rank the tasks based on their difficulty level. Fig. \ref{fig:result_difficulty} shows their rankings based on the patterns used in the tasks. Pattern A was used only in Task 0 when they worked without the robot.
Typically, the perceived difficulty of the task depends on the challenge of memorizing the first pattern within 45 seconds and then recalling it while arranging the blocks, with the second given pattern as a hint. The second patterns, Patterns A\textsubscript{2}, B\textsubscript{2}, C\textsubscript{2}, and D\textsubscript{2}, had 9, 12, 6, and 9 partially known spots, respectively. However, the number of these unknown spots is not the sole factor in determining their difficulty level; the complexity of memorizing them also plays a crucial role. This is why most participants considered Pattern D\textsubscript{2} with 9 unknown spots simpler than Pattern C\textsubscript{2} with 6 unknown spots because it was more challenging for them to memorize Pattern C\textsubscript{1} compared to Pattern D\textsubscript{1}. Participants also found Pattern B to be one of the most challenging patterns, as expected.

\begin{figure}
    \centering
    \includegraphics[width=\linewidth]{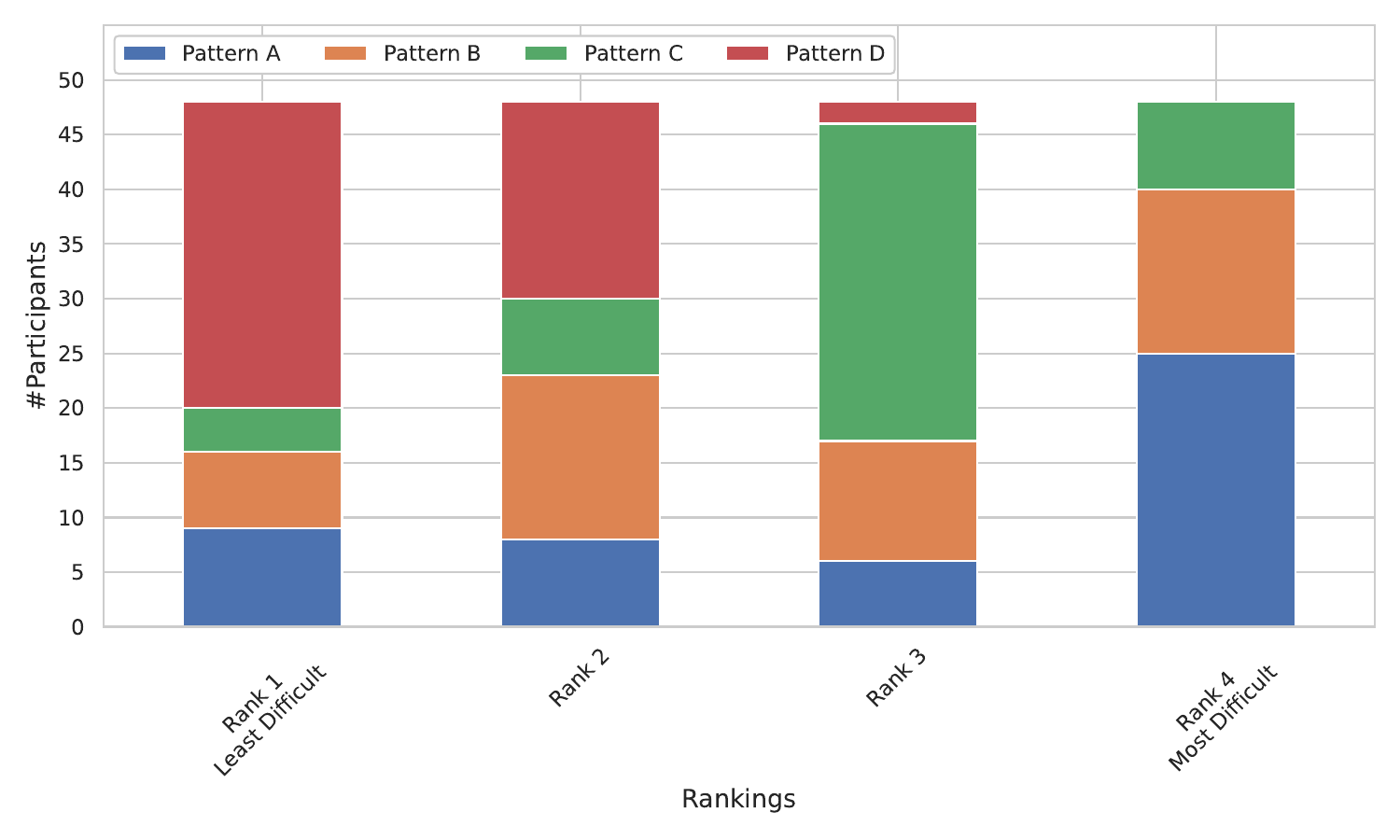}
    \caption{Ranking of task difficulty by participants}
    \label{fig:result_difficulty}
\end{figure}

\item \textbf{Self-confidence: }We have also analyzed participants' self-confidence based on the patterns, revealing a significant difference ($H(3) = 24.42$, $p\ll .001$). Pattern A was consistently used for Task 0 when participants worked alone. Fig.~\ref{fig:result-confidence_pattern} illustrates that participants' self-confidence in accomplishing Pattern 0 (Task 0) is significantly lower than that for Patterns 2 and 3 (no significant difference observed between Patterns 0 and 1). Furthermore, we observed that participants exhibited significantly higher self-confidence in performing Pattern 3 compared to Patterns 1 and 2 (no significant difference was noted between Patterns 1 and 2).

\item \textbf{Relative Trust: } A significant difference was found when evaluating participants' relative trust for the patterns ($H(2) = 12.9$, $p=.001$). Fig.~\ref{fig:result_relative_trust} shows participants' relative trust based on the patterns, where we can observe a significant difference between Patterns B and D as well as Patterns C and D.

\item\textbf{Perceived mental demand: } In Fig.~\ref{fig:result_mental_pattern}, as Pattern A was used only for Task 0, we only considered Patterns B, C, and D, revealing a significant difference ($H(2) = 8.05$, $p=.019$). The results show that participants perceived Pattern D as less mentally demanding than Patterns B and C.

\item \textbf{Perceived effort: } Additionally, in Fig.~\ref{fig:result_effort_pattern}, participants perceived significantly higher effort for completing Pattern B compared to Patterns C and D.
\end{itemize}

\textbf{Discussion: }
The subjective measurements support Hypothesis~\ref{hypo1a}, suggesting that participants' trust in the robot grows over time during the sessions. Additionally, according to the results, participants' self-confidence improved during the sessions as they collaborated more with the robot. This supports Hypothesis~\ref{hypo2A}.
Participants' answers also indicate that they expected more help from the robot during the sessions, supporting Hypothesis~\ref{hypo3}.
The results also partially support Hypothesis \ref{hypo4}. As discussed, participants perceived less workload in terms of mental demand, performance, effort, and frustration, supporting Hypothesis \ref{hypo4}. Contrary to Hypothesis \ref{hypo4}, participants also perceived less temporal demand when collaborating with the robot. However, participants' perceived physical demand did not significantly change when they completed the task in collaboration with the robot.
\emph{To summarize, the most important conclusion drawn from this study is that the collaboration could enhance participants' perception of the robot and collaboration and decrease their perceived workload.}

We also evaluated the participants' self-confidence, relative trust, and workload (mental demand and effort) based on the patterns. The results show that, as planned, Pattern B was more challenging for the participants than Patterns C and D. This also supports Hypothesis~\ref{hypo2B}, which states that participants will trust the robot more than their own abilities for challenging tasks.

\subsection{Participants' Actions and Performance}

\textbf{Overall preference: } In the online phase, we interviewed participants and asked about their strategies and preferences for collaborating with the robot. Probing their inclination toward leading or following, we inquired whether they preferred assigning tasks to the robot or being assigned tasks. Through detailed discussions, we discerned distinct preferences, categorizing participants into four groups: lead, collaborative-lead, collaborative-follow, and follow. The groups ``lead" and ``follow" represent participants with completely leading or following preferences, and the groups ``collaborative-lead" and ``collaborative-follow" represent participants who preferred to collaborate while having a slight preference to lead or follow, respectively. Four participants fell under neither of these four groups. One of them preferred to neither lead nor follow, but do his own tasks, and he did not assign any subtask to Fetch and did all subtasks assigned by Fetch. Three other participants quickly began doing subtasks because they did not want to forget the pattern. They mostly chose subtasks for themselves but were also willing to do subtasks assigned by Fetch. Table~\ref{tab:real_preferenc} shows the number of participants for each group. This indicates that 37  out of 48 participants had more of a leading preference, supporting Hypothesis \ref{hypo5}. 

\textbf{Task difficulty:}
At the end of Task 0, we counted the number of misplaced blocks on the shared table. We also recorded the number of wrong actions by participants, including placing or assigning wrong blocks and returning a correct block from the shared table during the experiment. 
Regardless of the number of errors, 28, 12, 9, and 3 participants made at least a mistake in respectively Patterns A, B, C, and D. Pattern A was used only in Task 0, with the most participants who made mistakes.

\begin{figure}
    \centering
    \includegraphics[width=0.8\linewidth]{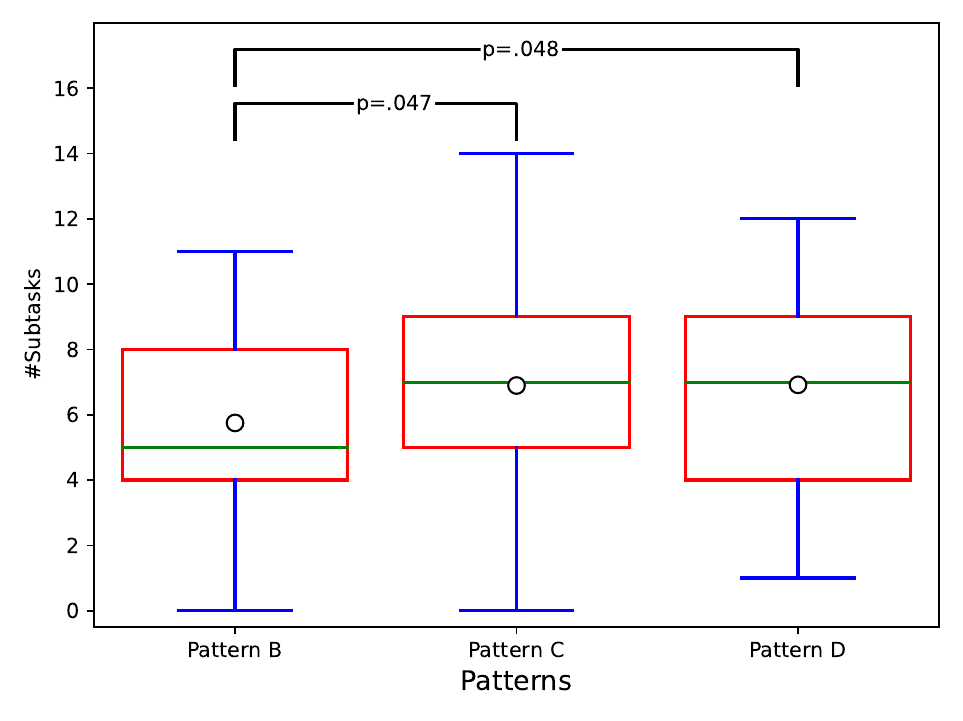}
    \caption{Number of subtasks assigned by participants to themselves in each task}
    \label{fig:result_human_assign_itself}
\end{figure}

\deleted{Additionally, we counted the number of subtasks that participants assigned to themselves based on the patterns. As depicted in Fig.~\ref{fig:result_human_assign_itself}, participants selected significantly fewer tasks for themselves in Pattern B compared to Patterns C and D. This indicates that they allowed the robot to assign more tasks to them. This finding aligns with their perception of task difficulty discussed earlier.}

\textbf{Task assignment: } 

\added{We counted the number of subtasks that participants assigned to themselves based on the patterns. As depicted in Fig.~\ref{fig:result_human_assign_itself}, participants selected significantly fewer tasks for themselves in Pattern B compared to Patterns C and D. This indicates that they allowed the robot to assign more tasks to them. This finding aligns with their perception of task difficulty discussed earlier.}

\begin{figure}
    \centering
    \includegraphics[width=0.8\linewidth]{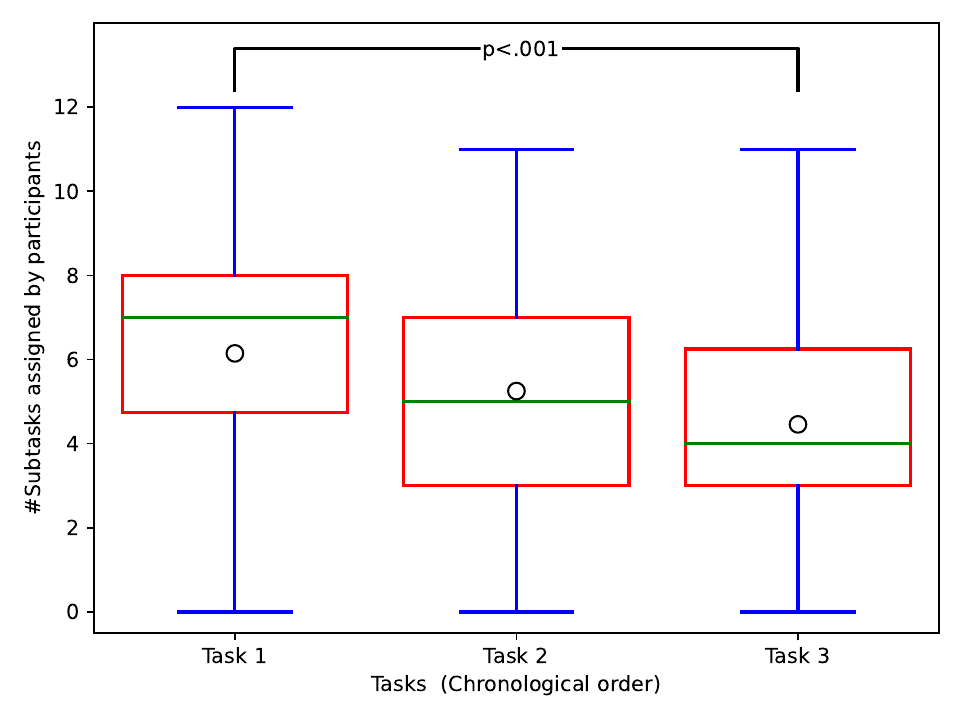}
    \caption{Number of blocks assigned by participants to Fetch in Tasks 1, 2, and 3}
    \label{fig:result_human_assign}
\end{figure}

Additionally, we measured the number of subtasks that participants assigned to the robot, revealing a significant difference between the tasks ($H(2) = 11.56$, $p=.003$).
Fig.~\ref{fig:result_human_assign} shows that participants preferred to assign fewer blocks to the robot, and there is a significant difference between Tasks 1 and 3. 

\textbf{Discussion: }
The results of the interviews show that most of the participants preferred to take on the leading role and have more control over the robot, which aligns with Hypothesis~\ref{hypo5}. In addition, according to the interviews, participants found the robot to be slower than themselves and preferred to handle more blocks. However, this preference may change if they were faced with a longer task requiring more physical effort. Some participants also pointed out that if the task were longer or the blocks were further apart, they would choose to assign tasks to the robot, even at the cost of completion time, to reduce their physical effort. Furthermore, the higher number of participants who made at least one mistake in Task 0 (Pattern A) compared to the other tasks indicates that the robot could assist them in improving their performance and making fewer errors. Except for Pattern A, which was only used in Task 0, participants made the most mistakes in Pattern B, a fact reflected in the subjective analysis of their perception of the task difficulty. This led them to rely more on the robot's decisions and help.

\begin{table*}[!t]
    \centering
    \caption{Participants' preference based on interviews.}
    \label{tab:real_preferenc}
    \begin{tabular}{c|c|c|c|c|c}
    \toprule
        \textbf{Preference} & Lead & Collaborative-lead & Collaborative-follow & Follow & Other \\ \midrule
          \textbf{\# Participants} & 17 & 20 & 4 & 3 & 4\\ 
         \bottomrule
    \end{tabular}

\end{table*}

\subsection{Leadership and Followership style}
The analysis of followership styles reveals that the majority of participants (37 out of 48) are classified as ``exemplary" followers, and ten of them fall into the ``pragmatist" category. Only one participant exhibited traits of an ``alienated" follower, and we did not encounter any instances of passive or conformist followers among the participants. This observation can likely be attributed to their shared academic background, which places a strong emphasis on critical thinking and active engagement.

It is worth highlighting that all four participants listed in Table~\ref{tab:real_preferenc}, who neither assumed leadership nor followed the robot, are categorized as ``pragmatist" followers. However, due to the limited number of these observations, no definitive conclusion can be drawn, and more investigation is needed.

\begin{figure}
    \centering
    \includegraphics[width=\linewidth]{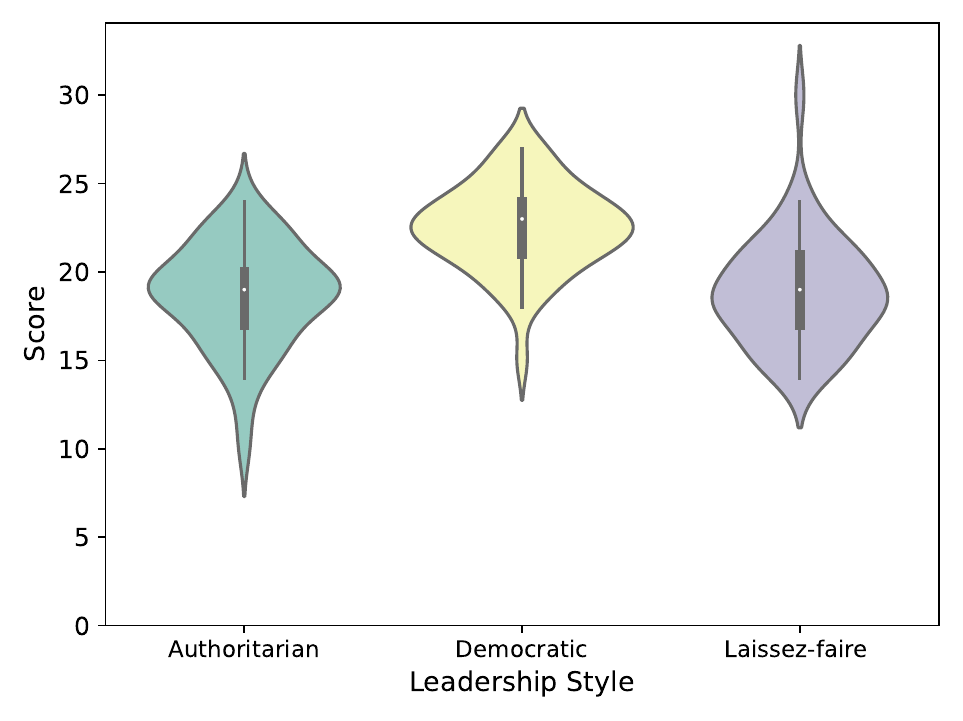}
    \caption{\replaced{Distribution of participants' leadership styles based on their responses to the leadership style questionnaire}{Caption}}
    \label{fig:result_leaderstyle}
\end{figure}

The distribution of leadership styles is depicted in Fig.~\ref{fig:result_leaderstyle}. 
The majority of participants exhibited a predominant ``democratic leadership" style. Among the four participants with no clear inclination toward leading or following, three displayed a dominant ``laissez-faire leadership" style. However, we cannot make conclusive statements based on this limited sample size. The sole participants characterized by an alienated followership style also demonstrated a notably dominant ``laissez-faire leadership" style and exhibited a leading-collaborative approach during the collaborative interactions.

We conducted an analysis of the correlation between engagement, critical thinking, and leadership styles in relation to various participants' factors. Intriguingly, our findings revealed a positive (negative) correlation between the number of tasks assigned by the robot to participants (the number of self-assigned tasks by participants) and their authoritarian leadership style. In other words, participants with authoritarian leadership tendencies were more inclined to accept additional subtasks assigned by the robot while selecting fewer tasks for themselves. Additionally, a moderate positive correlation was identified between participants' initial trust in the robot and their authoritarian leadership style.

\begin{figure*}
    \centering
    \includegraphics[width=0.99\linewidth]{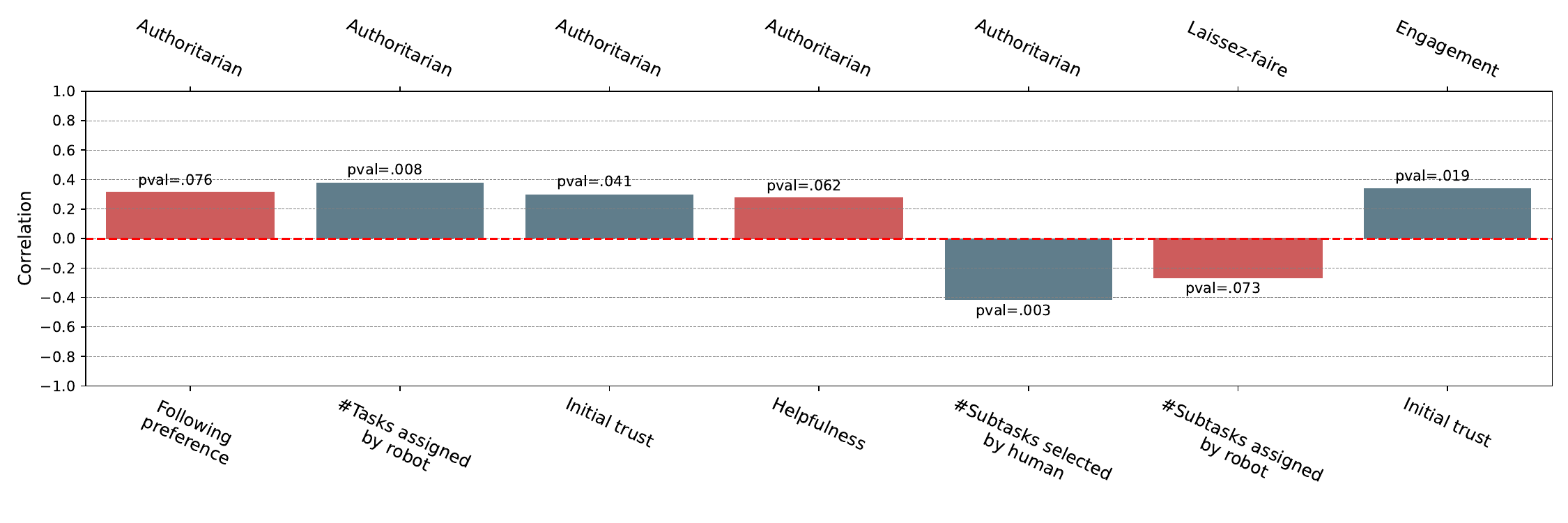}
    \caption{Correlations related to participants' leadership/followership styles}
    \label{fig:result_style_correl}
\end{figure*}

A positive, albeit nonsignificant, correlation was observed between these participants' following preference and their perceived helpfulness of the robot in relation to their authoritarian leadership style. However, this connection warrants further investigation in future research. We also encountered a nonsignificant negative correlation between participants' laissez-faire leadership style and the number of tasks assigned by the robot. Moreover, there was a positive correlation between participants' initial trust in the robot and their level of engagement. These correlations are shown in Fig.~\ref{fig:result_style_correl}. 

\section{Conclusion, limitations and future work}
Our study explored the potential enhancement of human-robot collaboration efficiency through proactive task planning and allocation. Unlike previous literature, which often neglects either human agents' leading/following preferences or their performance, we centered on achieving a balance between human agent preference and performance, while ensuring a high level of collaboration and positive human perception of the robot.

Trust is an integral part of this collaboration. We assessed how the robot's adaptive and proactive planning can improve participants' trust in the robot. The subjective measurement of participants' trust in the robot confirmed our Hypothesis~\ref{hypo1a}, showing trust enhancement as they collaborated more with the robot. Additionally, we hypothesized that participants' self-confidence would improve over collaboration with the robot, which was confirmed by analyzing their self-reported self-confidence (Hypothesis \ref{hypo2A}). Furthermore, we observed an increase in the expected helpfulness of the robot by participants, supporting Hypothesis~\ref{hypo3}. According to participants' answers to the questionnaires, they found the robot to be a teammate with positive traits and a strong working alliance, with whom they were willing to collaborate again in the future.

In support of Hypothesis~\ref{hypo2B}, the results indicated that for more difficult tasks, participants trusted the robot more than their own abilities, which led them to take relatively more following roles. Our proposed task planning method properly inferred this need and provided more help to participants by taking on more leading roles. The robot could also identify when participants struggled to remember the correct patterns and made errors, and accordingly, it fixed their errors and provided more help. This ability of the robot to adapt its task planning led us to hypothesize that participants would perceive, overall, a lower workload (Hypothesis~\ref{hypo4}). The results partially supported our hypothesis, and we observed an enhancement in participants' perceived mental demand, performance, effort, frustration, and unexpectedly temporal demand. However, contrary to our hypothesis, participants did not perceive a lower physical demand, likely due to the designed scenario focusing more on mental ability than physical abilities.

Based on interviews with participants, we categorized them into four groups: `lead', `collaborative-lead', `collaborative-follow', and `follow', with the majority falling into the first two categories. This supports Hypothesis~\ref{hypo5} that participants, while having a high level of trust in the robot, would prefer to take on more of a leading role and have more control over the collaboration. This finding can guide the design of collaborative scenarios and collaborative robots. 

Additionally, the results showed that participants selected fewer tasks for themselves and were allocated more subtasks in tasks that were more difficult since they preferred to follow the robot or made many errors, causing the robot to take back the leading role.

We also inquired about participants' leadership and followership styles using questionnaires. Our observations revealed that individuals with an authoritarian leadership style exhibited greater trust in the robot. They also allowed the robot to assign more subtasks to them. Additionally, we identified a positive correlation between their initial trust and the level of engagement. Nevertheless, a more comprehensive investigation involving a diverse participant group is warranted to delve deeper into this subject.

\subsection{\added{Takeaway Messages}}
\added{
The task planning framework and algorithm employed in this paper provide a foundational architecture that, with customization to suit specific contexts and tasks, can be applied to various tasks and collaborative scenarios. We tested it on the designed collaborative scenarios in a simulation environment and an actual robot in a user study. Although the results discussed are based on these particular studies, we can draw some general takeaway messages from them:}
\begin{itemize}

    \item \added{Efficiently designed adaptive planning, achieved by customizing the proposed task planning framework, can significantly enhance participants' perception of the robot and collaboration (e.g., satisfaction and trust in the robot).}
              
    \item \added{Adaptive planning improves team performance and reduces participants' perceived task load. This finding underscores the importance of adaptive strategies in enhancing overall collaboration efficiency.}
              
    \item \added{In the design of future collaborative scenarios, it is crucial to consider that humans typically prefer to maintain a degree of control when collaborating with machines or robots. Understanding and accommodating this preference can enhance the acceptance and effectiveness of human-robot collaborative systems.}
              
    \item \added{Human agents' preferences and performance can vary based on factors such as task difficulty. The adaptive planner demonstrates the ability to infer these changes and adjust the robot's plan accordingly, optimizing collaboration outcomes.}
              
    \item \added{Acknowledging the inevitability of human error in collaboration, it is essential to integrate error detection mechanisms into robot planning. The proposed adaptive task planning approach effectively detects errors and adjusts the robot's plan to maintain high team performance levels, even in the face of human errors.}
\end{itemize}

\subsection{Limitations and Future Work}
This study has certain limitations concerning its design and methods. All of our participants were recruited from the University of Waterloo campus and consisted mainly of young adults. However, our ultimate goal is to study working adults in settings such as manufacturing and warehouses. It is conceivable that these two groups hold significantly different expectations and perceptions of a robot teammate. To address this, involving actual working adults in such settings could enhance the practicality of our collaborative scenarios and the robots used. Despite efforts to simulate a working environment by employing an autonomous robot for pick-and-place tasks and creating a setup with features like a conveyor belt, safety equipment, and a graphical user interface, there is room for future enhancements to make the scenarios more closely resemble realistic environments, such as warehouse automation or assembly settings. Moreover, prevalent challenges in manufacturing settings, like sudden changes or unpredictable events typically managed by humans, could be incorporated to better align the study with real-life situations.

In our current study, we assumed the infallibility of the robot's decisions. However, real-world scenarios may involve instances where either the robot or the human \replaced{agent}{agents} makes errors, potentially without awareness, and believe their decisions are correct, which can lead to a conflict between two agents. Additionally, our study gauged the human agent's accuracy in selecting block colors as a measure of correctness, a metric easily comprehensible to participants. While practical measures like completion time or travel distance could be introduced, these might not be as easily understandable and measurable for human agents, potentially leading to conflicts. Numerous other avenues for exploration exist, including a comprehensive investigation into how human leadership and followership styles influence collaboration with a robot.

 \bibliographystyle{elsarticle-num} 
 \bibliography{ref}

\end{document}